\documentclass[a4paper,fleqn]{cas-dc}

\usepackage{amsmath,amsfonts}
\usepackage{algorithmic}
\usepackage{enumitem}
\usepackage{xcolor} % 支持颜色
\usepackage{graphicx}
\usepackage[justification=centering]{caption}
\usepackage{subcaption} 
\usepackage[numbers]{natbib}
\definecolor{myblue}{HTML}{1E90FF}  % 定义一个叫 
\definecolor{myred}{HTML}{f61e68}

\usepackage[linesnumbered,ruled,vlined]{algorithm2e}
\def\tsc#1{\csdef{#1}{\textsc{\lowercase{#1}}\xspace}}
\tsc{WGM}
\tsc{QE}
\tsc{EP}
\tsc{PMS}
\tsc{BEC}
\tsc{DE}
\begin{document}
\let\WriteBookmarks\relax
\def\floatpagepagefraction{1}
\def\textpagefraction{.001}

\shorttitle{Dynamic graph partition and task scheduling for GNNs computing in edge network}

\shortauthors{Wenjing X et~al.}

\title [mode = title]{GraphEdge: Dynamic Graph Partition and Task Scheduling for GNNs Computing in Edge Network}                      
% \tnotemark[1,2]

% \tnotetext[1]{This document is the results of the research
%    project funded by the National Science Foundation.}

% \tnotetext[2]{The second title footnote which is a longer text matter
%    to fill through the whole text width and overflow into
%    another line in the footnotes area of the first page.}

\author[1,2]{Wenjing Xiao}[style=chinese, type=editor,
                        % auid=000,bioid=1,
                        % prefix=Sir,
                        % role=Researcher,
                        % orcid=0000-0001-7511-2910
]
% \cormark[1]
\ead{wenjingx@gxu.cn}
% \ead[url]{www.cvr.cc, cvr@sayahna.org}
% \credit{Conceptualization of this study, Methodology, Software}

\author[1,2]{Chenglong Shi}[style=chinese]
\ead{chenglongs@st.gxu.edu.cn}

\author[1,2]{Miaojiang Chen}[style=chinese]
\ead{mjchen_cs@gxu.edu.cn}
\cormark[1]

% 4 author
\author[5]{Zhiquan Liu}[style=chinese]
\ead{zqliu@vip.qq.com}

\author[3,4]{Min Chen}[style=chinese]
\ead{minchen@ieee.org}

\author[5]{H. Herbert Song}[style=chinese]
\ead{h.song@ieee.org}

% Address/affiliation[1]
\affiliation[1]{organization={Guangxi Key Laboratory of Multimedia Communications and Network Technology}, city={Nanning, 530004}, country = {China}}
 \affiliation[2]{organization={School of Computer, Electronics and Information, Guangxi University, Nanning, 530004}, country = {China}}   
% Address/affiliation[2]
\affiliation[3]{organization={School of Computer Science and Engineering, South China University of Technology, Guangzhou 510006}, country = {China}}
% Address/affiliation[2]
\affiliation[4]{organization={Pazhou Laboratory, Guangzhou 510330}, country = {China}}
% Address/affiliation[2]
\affiliation[5]{organization={College of Cyber Security, Jinan University, Guangzhou 510632}, country = {China}}
\affiliation[6]{organization={Department of Information Systems, University of Maryland, Baltimore County (UMBC), Baltimore, MD 21250}, country = {USA}}

\cortext[cor1]{Corresponding author}
% \cortext[cor2]{Principal corresponding author}

% \fntext[fn1]{This is the first author footnote. but is common to third
% \fntext[fn2]{Another author footnote, this is a very long footnote and
%   it should be a really long footnote. But this footnote is not yet
%   sufficiently long enough to make two lines of footnote text.}

% \nonumnote{This note has no numbers. In this work we demonstrate $a_b$
%   the formation Y\_1 of a new type of polariton on the interface
%   between a cuprous oxide slab and a polystyrene micro-sphere placed

\begin{abstract}
% This template helps you to create a properly formatted \LaTeX\ manuscript.
With the exponential growth of Internet of Things (IoT) devices, edge computing (EC) is gradually playing an important role in providing cost-effective services. However, existing approaches struggle to perform well in graph-structured scenarios where user data is correlated, such as traffic flow prediction and social relationship recommender systems. In particular, graph neural network (GNN)-based approaches lead to expensive server communication cost. To address this problem, we propose GraphEdge, an efficient GNN-based EC architecture. It considers the EC system of GNN tasks, where there are associations between users and it needs to take into account the task data of its neighbors when processing the tasks of a user. Specifically, the architecture first perceives the user topology and represents their data associations as a graph layout at each time step. Then the graph layout is optimized by calling our proposed hierarchical traversal graph cut algorithm (HiCut), which cuts the graph layout into multiple weakly associated subgraphs based on the aggregation characteristics of GNN, and the communication cost between different subgraphs during GNN inference is minimized. Finally, based on the optimized graph layout, our proposed deep reinforcement learning (DRL) based graph offloading algorithm (DRLGO) is executed to obtain the optimal offloading strategy for the tasks of users, the offloading strategy is subgraph-based, it tries to offload user tasks in a subgraph to the same edge server as possible while minimizing the task processing time and energy consumption of the EC system. Experimental results show the good effectiveness and dynamic adaptation of our proposed architecture and it also performs well even in dynamic scenarios.
% \noindent\texttt{\textbackslash begin{abstract}} \dots 
% \texttt{\textbackslash end{abstract}} and
% \verb+\begin{keyword}+ \verb+...+ \verb+\end{keyword}+ 
% which
% contain the abstract and keywords respectively. 

% \noindent Each keyword shall be separated by a \verb+\sep+ command.
\end{abstract}

% \includegraphics{figs/grabs.pdf}

% 不需要的附加的第一页
% \begin{highlights}
% \item Research highlights item 1
% \item Research highlights item 2
% \item Research highlights item 3
% \end{highlights}

% Keywords

\maketitle

\section{Introduction}
\label{Introduction}
With the rapid development of 5G, IoT, and artificial intelligence (AI) technologies, the era of the Internet of Everything has arrived~\cite{SecureIoT2024},~\cite{10818776},~\cite{luo2023enabling},~\cite{luo2024ripple},~\cite{luo2024edge}. This also derives edge computing (EC) scenarios with graph-structure features such as social graphs~\cite{li2018influence} and multi-sensor environment-aware networks~\cite{PPRU2023},~\cite{EViT24}. In such EC scenarios, users are related to each other and no longer independent individuals, and their computing tasks usually require access to the corresponding data of their neighboring users. For example, in multi-sensor environment-aware networks, we can model the scenario as a graph, where the sensing devices are treated as users and represented as vertices, the communication and collaboration states between users are represented as edges. In this way, the vertices and edges in the graph represent the users' data and their data associations. When certain user device performs computing tasks of environment prediction, its association relationship with other user devices and correlation of data need to be considered. Since the relationship between users cannot be measured using the traditional Euclidean distance metric, the graph-structured EC scenarios are non-Euclidean. However, existing research on EC~\cite{han2024novel},~\cite{tang2025real},~\cite{rai2024agricultural},~\cite{wan2024pflow} mainly focus on optimizing CNNs-based computing tasks on traditional scenarios with Euclidean data~\cite{myagmar2023survey}, lacking the consideration of the association between users and the graph topology formed by the users. To this end, it is crucial to investigate novel task offloading strategies for efficiently handling tasks with non-Euclidean data~\cite{hasan2024mm}.

Recently, Graph Neural Networks (GNNs) map
complex relationships between nodes into a high-dimensional
feature space to fully learn their intrinsic dependencies, and have shown significant advantages in processing non-Euclidean graph data. In the EC system for GNN tasks, the complete and identical GNN models are deployed on edge servers located in different geographic locations.
Each user as a graph vertex offloads its task to communicable edge servers, and next the offloaded task data will be inferred by the GNN model in the form of graph data. 
Then GNN aggregates the feature data of associated users using the graph structure and linearly changes the aggregation results by applying them to the weight matrix. Fig.~\ref{scenario} illustrates an edge computing system for GNN tasks. It can be seen that the data of different users are often offloaded to different edge servers. When a certain edge server performs the GNN aggregation process, it often needs to obtain the data of associated users on other edge servers. This process is called \textit{message passing}~\cite{ju2024cool} and results in massive data transmission among edge servers. 

However, existing researches only focus on designing cost-efficient task scheduling schemes and fail to the consideration of optimizing the graph layout~\cite{kumar2023eeoa},~\cite{ali2022cost}. 
% In a word
Furthermore, these methods still do not consider the following three aspects: (1) \textit{Associations among Graph Vertices}. 
% \textbf{Firstly}, 
The graph layout formed by user vertices will significantly affect the inference cost of 
%ECs 
EC system for GNN tasks. If the weak-association task data are placed centralized in the same edge servers and strong-association task data are placed in different edge servers, it will incur large data transfer among edge servers during GNN inference and lead to a lot of inference cost. Note that the placement is the task offloading of the graph vertex data corresponding to users. Furthermore, due to the irregular relationships between graph vertices, it is challenging to rationally optimize the graph layout according to the aggregation characteristics of GNN. (2) \textit{Dynamics of Graph Topologies}. 
% \textbf{Secondly}, 
The states of users participating in GNN computational tasks often dynamically change over time, 
which leads to changes in the graph layout formed by the user's topology.
Such frequent changes in graph layout often result in the significant degradation of task execution efficiency and resource utilization. Thus, it is required that the EC system must be able to perceive the changes in the number and relationships of users. (3) \textit{Real-time Optimal Offloading}. 
% \textbf{Thirdly}, 
When the graph layout of users has been updated, the offloading scheme of the previous is often no longer optimal. If it is still used, the inference performance even becomes inferior. To obtain the most efficient offload decision, the offload algorithm needs to re-perform the whole solving process for the new graph layout, which incurs large computation and time cost. Thus, the offloading scheme of GNN tasks needs to adapt to the dynamic changes in graph-structured EC scenarios.

To address the above issues in graph-structured EC scenarios, this paper proposes GraphEdge, an efficient GNN-based edge computing architecture. It can dynamically aware the graph layout of EC scenarios for graph layout optimization and task offloading. Specifically, firstly, we propose the dynamic graph model to represent changes in users and their association states. This model adopts a \textit{mask} module and position attributes to maintain the user's states and changes. Secondly, based on this graph model, we propose a hierarchical traversal graph cut algorithm to optimize the original graph layout into a new graph layout consisting of multiple subgraphs. In this new graph layout, user vertices in same subgraphs are strongly associated, and user vertices between subgraphs are weekly associated. Moreover, task offloading based on the new graph layout can reduce the \textit{message passing} cost during GNN inference. At this point, since the offloading strategy determines which server the tasks in each subgraph are offloaded to, it is termed graph offloading. 
Thirdly, we propose a deep reinforcement learning-based graph offloading algorithm, termed DRLGO, which is real-time aware of the changes in graph applications and EC networks to efficiently generate optimal graph offloading decision based on optimized graph layout. Overall, our contributions are as follows:

\begin{figure}[t]
    \centering
    \includegraphics[width=8.3cm]{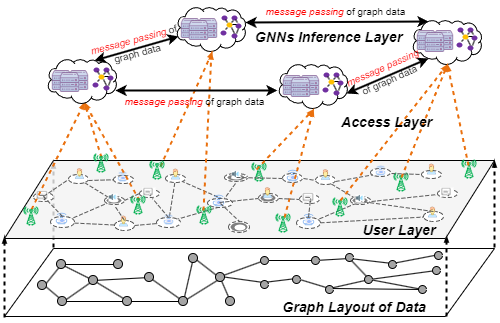}
    \caption{EC system for GNN tasks.}
    \label{scenario}
\end{figure}

\begin{itemize}[itemsep=-0.3em, topsep=2pt] 
    \item  We propose an efficient GNN-based edge computing architecture, which transmits graph data to different edge servers for GNN inference, achieving GNN processing over multi-edge cloud collaboration.
    \item We propose a hierarchical traversal graph cut algorithm that cuts the whole graph for users into series subgraphs. 
    The vertices in the same subgraph are strongly associated while different subgraphs are weakly associated. 
    And the cross-server communication cost caused by message passing between subgraphs can be minimized in GNN inference.
    \item We propose the DRL-based graph offloading algorithm to efficiently generate optimal graph offloading decisions in dynamic graph-structured EC scenario, it considers the heterogeneity of servers and can minimize the system cost to process user tasks.
    \item Extensive simulation experiments are conducted to verify the effectiveness and superiority of our proposed method in the graph-structure EC system. Experimental Results demonstrate that DRLGO can minimize system cost even in the dynamic EC scenario. Moreover, HiCut can effectively reduce cross-server communication cost during GNN inference. 
\end{itemize}

The rest of this paper is organized as follows: Section~\ref{Realted works} briefly reviews the previous works in EC, GNN, GNN and DRL applications in EC. Section~\ref{System model} will give the system model and optimization objective of our work. The proposed graph cut algorithm, HiCut will be introduced in Section~\ref{HiCut}. Next, we'll describe the proposed graph offloading algorithm, DRLGO in Section~\ref{DRLGO}. And then, the results and analysis of the experiment will be given in Section~\ref{Experiment}. At last, We will give a conclusion of our work in Section~\ref{Conclusion}.

% The Elsevier cas-dc class is based on the
% standard article class and supports almost all of the functionality of
% that class. In addition, it features commands and options to format the
% \begin{itemize} \item document style \item baselineskip \item front
% matter \item keywords and MSC codes \item theorems, definitions and
% proofs \item lables of enumerations \item citation style and labeling.
% \end{itemize}

% This class depends on the following packages
% for its proper functioning:

% \begin{enumerate}
% \itemsep=0pt
% \item {natbib.sty} for citation processing;
% \item {geometry.sty} for margin settings;
% \item {fleqn.clo} for left aligned equations;
% \item {graphicx.sty} for graphics inclusion;
% \item {hyperref.sty} optional packages if hyperlinking is
%   required in the document;
% \end{enumerate}  

% All the above packages are part of any
% standard \LaTeX{} installation.
% Therefore, the users need not be
% bothered about downloading any extra packages.

\section{Realted works}
\label{Realted works}
In this section, we first introduce the related works in EC and GNN. Then, we introduce the applications of GNN and DRL in edge computing.
\subsection{Edge computing}
EC is a distributed service architecture, it sinks central cloud service capabilities to edge servers that are closer to users. In recent years, the rapid development of IoT has facilitated researchers to conduct many studies on EC. These researches mainly focus on EC scenarios and task offloading optimizations~\cite{xiao2024adaptive}. Dhuheir et al.~\cite{dhuheir2022deep} classified the EC scenario into hotspot and non-hotspot areas, and solved the problem of automatic monitoring and path planning for Unmanned Aerial Vehicles (UAVs) in their research scenario. Qian et al.~\cite{qian2022joint} investigated the task offloading optimization problem in the Maritime Internet of Things (M-IoT). 
Hao et al. investigated the task offloading problem in the DT-enabled URLLC mobile edge network and proposed a more effective DT-assisted robust task offloading scheme based on learning consisting of decision and deviation networks~\cite{hao2023digital}.
In~\cite{hao2023joint},~\cite{seid2021collaborative},~\cite{wei2023joint}, DRL was used to address collaborative computational offloading and resource allocation in EC networks. 
Chen et al.~\cite{chen2023sgpl} used a game-theoretic approach to improve the anti-jamming ability between smart gaming devices and proposed a novel anti-jamming cooperative computational model for smart games.

\subsection{Graph neural network}
Graph neural networks are mainly used to process graph structure data contains vertices and edges generated by these vertices' associations. At the same time, each vertex usually has its own feature data~\cite{zhao2024model}. The reason why GNNs can learn the information of neighborhoods using associations between vertices is because they combine the graph embedding techniques~\cite{ahmed2024enhancement} and the neural network operator techniques~\cite{costarelli2015neural}. The inference process of GNN is described in Section~\ref{Introduction} and will not be repeated here. Moreover, the vertices in the graph can get information about each other after a number of \textit{message passing} iterations~\cite{bruna2013spectral}. All this presupposes that the work~\cite{hamilton2017inductive} implements the convolutional techniques in graphs which optimize the aggregation operation of GNN and make the learning of graphs possible.
%~\cite{saensouk2022curcuma}
 
Taking GCN \cite{ma2024spatio} as an example, the formula for the aggregation and update of GCN is as follows (considering the inference of a GCN with two layers), The input to the GCN is a graph $\mathcal{G}=(V, E)$, where $V$ denotes the set of vertices in the graph and $E$ denotes the set of edges between vertices in the graph.
    \begin{equation}
       \hspace{2.5em} H^{(\kappa+1)}=\delta (\widetilde{D}^{-\frac{1}{2}}\widehat{A}\widetilde{D}^{-\frac{1}{2}}H^{(\kappa)}\phi^{(\kappa)} ),
        \label{eq1}
    \end{equation}
where $H^{(\kappa+1)}$ represents the hidden representation of the vertices in the $\kappa$th layer after aggregation, $\widehat{A}$ is the adjacency matrix with the addition of the self-loop, and $\widetilde{D}$ represents the matrix of the node degrees after the addition of the self-loop. $\delta$ is a nonlinear activation function. $\phi^{(\kappa)}$ represents the parameters of the $\kappa$th layer of GCN. The output of a two-layer GCN can be expressed as:

    \begin{equation}
        \hspace{1.1em}\Psi \left ({X,A}\right) =\delta   \left ({\widehat {A}\,ReLU\left ({\widehat {A}\, X\,W_{0}}\right)W_{1}}\right),
        \label{eq2}
    \end{equation}
where $X$ represents the feature data of the vertices. $W_{0}$ and $W_{1}$ denote the weight matrices for different layers of GCN, respectively.
\subsection{GNN applications in EC}
The application of GNN in EC implies the characteristics of distributed services due to the distributed character of EC. Much work does not take into account the limited resources and dynamic nature of the EC system for GNN tasks. Zeng et al.~\cite{zeng2022fograph} investigated the server-side distributed GNN processing system in edge services, but the cost statistics were not enough. This work only metrics such as network communication latency and throughput, and does not consider the overall energy and time cost of the system to process user tasks. He et al.~\cite{he2021fedgraphnn} studied distributed GNN systems under the federated learning paradigm, but this study was based more on security and did not take into account the performance of heterogeneous scenarios and limited resources. In works~\cite{tripathy2020reducing} and~\cite{zheng2020distdgl}, the GNN services are considered but EC scenarios are not dynamic. Zeng et al.~\cite{zeng2022gnn} studied GNN services in dynamic heterogeneous EC scenarios but did not consider the optimization of task offloading cost based on the GNN aggregation characteristics.

\subsection{DRL applications in EC}
 The applications of DRL in EC mainly focus on offloading strategy exploration for user tasks and path planning for mobile edge servers, such as unmanned vehicles and UAVs.
 Lakew et al.~\cite{lakew2022intelligent} investigated intelligent offloading and resource allocation problem in heterogeneous Air Access Internet of Things (AAIoT) networks, based on MADDPG, they explored the task offloading policy which minimized the total energy cost of the system. Peng et al.~\cite{peng2020multi} proposed a resource management scheme also based on MADDPG in UAV-assisted Internet of Vehicles (IoV). Seid et al.~\cite{seid2021collaborative} proposed an optimal task offloading scheme based on deep Q-learning (DQN) in autonomous driving scenarios, but the dynamic adaptation of the algorithm has not been explored. Furthermore, most of the work on combining GNN with DRL only uses GNN as an auxiliary tool for DRL. For example, in works \cite{sun2021graph},~\cite{li2023exploring},~\cite{li2024dynamic},~\cite{cao2024dependent},~\cite{wang2025graph}, etc. the researchers consider the dynamic EC scene as an acyclic graph and extract topological features of the scene using GNN. Then the DRL approach is used to explore the task offloading decision by combining the topological features.

 \subsection{Differences with existing work}
 Existing work considered static or limited dynamic EC scenarios. These scenarios are modeled as graphs, with edge and user devices as vertices and communication states or interpersonal relationships as edges. GNNs are typically used to extract vertex features, which are then processed by algorithms like DRL for task offloading or resource allocation. These works do not optimize the system energy consumption and task processing time for GNN aggregation characteristics. In short, the main differences between our work and existing work are as follows:

  1) We consider a more realistic dynamic scenario where user locations, numbers, and associations can change arbitrarily.
 
  2) The pre-trained GNN model is deployed on each edge server to process the received graph task data, rather than serving as part of the algorithm.
 
  3) We partition strongly correlated vertices into a subgraph based on GNN aggregation characteristics. For the partitioned subgraph, we use DRL to explore the optimal graph offloading strategy.

\section{System model}
\label{System model}
In this section, the system model will be introduced. We first introduce the execution process of our EC system. Then we describe the dynamic graph model, graph offloading communication model, and GNN computation model. Finally, we introduce the objective function and optimization task of our proposed problem.

\begin{figure}[h]
    \centering
    \includegraphics[width=8.3cm]{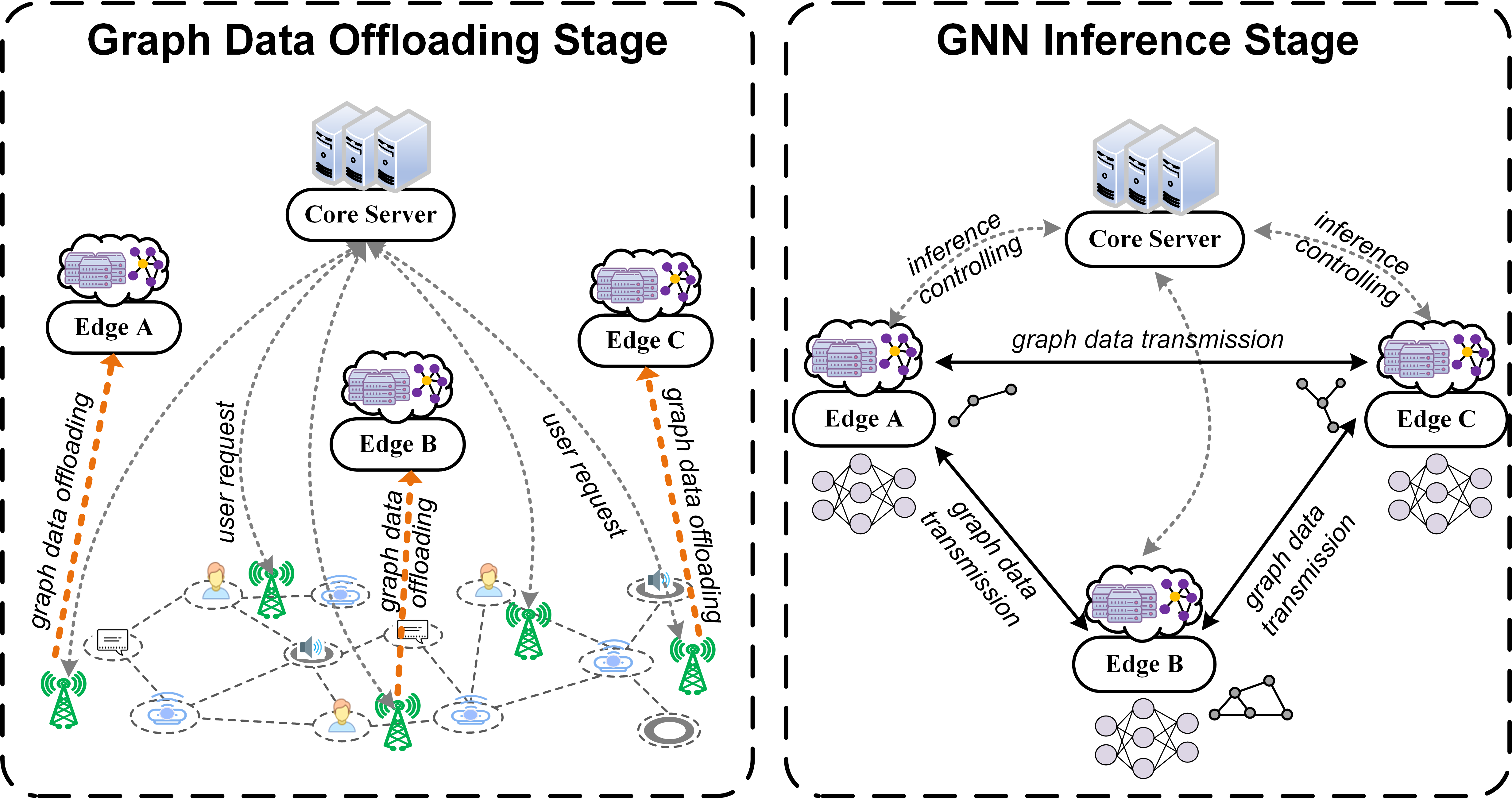}
    \caption{The processing flow of the GNN-based EC system.}
    \label{system_structure}
\end{figure}

\vspace{-1em}
\subsection{System overview}
\label{sec:sys-overview}
We consider the GNN-based edge computing architecture as shown in Fig.~\ref{scenario}. GNN models are deployed on edge servers located in different geographic locations. Users offload their tasks to the edge server through the access points (APs). Each edge server will receive the data of several users and take them as the input of the GNN model to execute inference. From the perspective of an edge server executing tasks, the data uploaded by each user is essentially a graph vertex for the input of GNN model at this edge server. Thus, the EC controller (core server) in our architecture first perceives the topology of users who participated in GNN computation and represents the association of user data as a graph. This graph is the original graph layout. Then the graph cut algorithm is adopted to optimize the original graph layout %representing EC 
into a new graph layout consisting of multiple subgraphs, minimizing the message passing between subgraphs when GNN is in inference. Based on the new graph layout, the EC controller generates the optimal graph offloading strategy, it decides which edge server each subgraph is offloaded to, and then broadcasts the offloading decision to each user. Subsequently, each user follows the offloading decision to upload the task data to edge servers for GNN computing. 
Eventually, the offloaded tasks will be used as the input for GNN inference. After GNN inference, each edge server will return the results to users. The processing flow of the proposed architecture is shown in Fig.~\ref{system_structure}.

\begin{table}[width=\linewidth,cols=2,pos=t]
\footnotesize
\caption{Main symbols definition.}
\label{table1}
\begin{tabular*}{\tblwidth}{@{} CC@{} }
\toprule
Notation & Description\\
\midrule
$\sigma$ & Noise power \\
$\omega$ & Edge network \\
$N$ & Number of users \\
$X_{i}$ & Task data size of $U_{i}$ \\
$\mathcal{G}_{sub}$ & Optimized graph layout \\
$|\mathcal{N}_{i}|$ & Number of neighbors of $U_{i}$ \\
$\vartheta$ & Unit data update cost in GNN  \\
$M$ & Number of APs and edge servers \\
$\mu$ & Unit data aggregation cost in GNN \\
$\mathcal{G}$ &  Original graph layout awared by EC system \\
$T_{all}$ & Time cost for system complete all user’s tasks \\
$I_{upd_{\kappa}}$ & Update cost at the $\kappa$th layer of GNN \\
$I_{agg_{\kappa}}$ & Aggregation cost at the $\kappa$th layer of GNN \\
$R_{i,m}$ & Uplink rate between $U_{i}$ and $AP_{m}$ \\
$T_{i,m}^{up}$ & Time cost for $U_{i}$ to upload task to $AP_{m}$ \\
$T_{i,f_{k}}^{com}$ & Time cost for GNN process the task data of $U_{i}$ \\
$I_{i,m}^{up}$ & Energy cost for $U_{i}$ to upload task to $AP_{m}$ \\
$I_{all}$ & Energy cost for system complete all user’s tasks \\
$R_{k,l}$ & Uplink rate between $SV_{k}$ and $SV_{l}$ \\
$B_{i,m}$ & Transmission bandwidth between $U_{i}$ and $AP_{m}$ \\
$B_{k,l}$ & Transmission bandwidth between $SV_{k}$ and $SV_{l}$ \\
$\varsigma_{i,m}$ & Energy cost of unit data uploaded from $U_{i}$ to $AP_{m}$ \\
$\varsigma_{k,l}$ & Energy cost of unit data transferred from $SV_{k}$ to $SV_{l}$ \\
$T_{k,l}^{tran}$ & Time cost for $SV_{k}$ to transfer data to $SV_{l}$ \\
$I_{k,l}^{com}$ & Energy cost for $SV_{k}$ to transfer data to $SV_{l}$  \\
$\varsigma_{k,l}$ & Energy cost of unit data transfered from $SV_{k}$ to $SV_{l}$ \\
$\phi$ & Energy cost for unit data to use activation function \\
$f_{k}$ & CPU cycles per second of GNN process unit data \\
\bottomrule
\end{tabular*}
\end{table}

As shown in Fig.~\ref{scenario}, our EC system mainly consists of three layers: the user layer, the access laye, and the GNN inference layer. The user layer contains $N$ users, represented as $U=\left \{ U_{n}|n=1,2,\dots,N \right \}$. In the remaining two layers, the number of APs and edge servers is the same and denoted by $M$, since each AP is associated with an edge server. Thus, the edge servers and APs can denote as $SV=\left \{ SV_{m}|m=1,2,\dots,M \right \} $ and $AP=\left \{ AP_{m}|m=1,2,\dots, M \right \} $ respectively. Let $e_{ij} \in \{0, 1\}$ to denote the association state between $U_{i}$ and $U_{j}$, where $e_{ij}=1$ indicates that the two are associated, and otherwise not associated. The communication state between edge servers is denoted by $\eta_{kl} \in \{0,1\}$, $\eta_{kl}=1$ indicating $SV_{k}$ and $SV_{l}$ are communicating, and the opposite indicating they are not in communication. The entire edge network adopts wireless channel communication and the channels between different APs do not interfere with each other. Besides, let $w_{ik} \in \{0, 1\}$ denotes the offloading decision variable indicating whether $U_{i}$ uploads its task data to $SV_{k}$. And if $w_{ik}$=1, it means upload, otherwise not upload. Table~\ref{table1} shows the main symbols used to describe our EC system.
\subsection{Dynamic graph model} 
Due to the dynamic states of users in the EC system, the graph layout formed by the users and the data associations between users will also change. To ensure that the graph layout perceived by the EC controller can represent the changes in data association, we propose the dynamic graph model to extend the graph layout perceived by the EC controller. Let $\mathcal{G}(t) = (V(t), E(t))$ denote the graph layout perceived by the EC controller at time step $t$. $V=\left \{ V_{n}|n=1,2,\dots,N \right \} $ is the set of vertices in $\mathcal{G}$, and each vertex corresponding to a user in the EC system. $E=\left \{ (V_{i},V_{j})|\forall i,j \in N , i \ne j \right \} $ is the set of edges in $\mathcal{G}$, representing the associations between the vertices in $\mathcal{G}$. In our EC system, users have three types of dynamic states: (1) The changes in users' location. (2) The changes in number of users. (3) The changes in user's associations. The following describes how to extend $\mathcal{G}$ based on the user's dynamic states. To represent user location changes, all vertices add the position attribute, which values synchronized to the corresponding user location. And the coordinates of $U_{i}$ at time step $t$ are given by $(x_i(t), y_i(t))$. To represent changes in the number of users, a $mask$ module is added to $\mathcal{G}$. The $mask$ is an array of length $N$ and the initial value of all elements is 1. If some users drop out or quit, the values of the corresponding vertices in the $mask$ will be updated to 0, and their associations with other vertices will be removed. When adding new users, positions in the $mask$ with a value of 0 will be set to 1 based on the number of new users. Meanwhile, new vertices will have the same position value and data associations with newly added users. For changes in the associations between users, it only needs to update the edges in $E$. In this way, we extend $\mathcal{G}$ to the dynamic graph model so that the states of users and data associations between them can be dynamically represented.
\subsection{Offloading communication cost}
The offloading communication model of the GNN-based EC system consists of three parts: graph offloading from users to edge servers, the data transmission among edge servers during the GNN inference, and the inference results transmission from edge servers to users. Since the data size of the inference result is too small, this paper mainly considers the cost of the first two and ignores the last one. The details are described in the following.

The channel gain between $U_i$ and $AP_{m}$ at time step $t$ is given by $h_{i,m}(t)$. $h_{i,m}(t)$ is modeled using the free space path loss model $h_{i,m}(t) = \varrho_{0}(d_{i,m}(t))^{-2}$, where $\varrho_{0}$ represents the channel gain at the reference distance $ d_{0} = 1$, and $d_{i,m}(t)$ denotes the distance between $U_{i}$ and $AP_m$ at time step $t$. In our EC system, the positions of APs and edge servers will not change after deployment and the communication quality between the AP and the edge server is good, so their communication cost can be ignored. Let $h_{0}$ be the channel gain between edge servers. 

First, we introduce the communication cost of the user offloading graph data to AP. At time step $t$, when $U_{i}$'s graph data is offloading to $AP_{m}$, the uplink rate between the two denoted as $R_{i,m}(t)$, it can be denoted as: 
\begin{equation}
    \hspace{1em}R_{i,m}(t)=B_{i,m}(t)\log_{2}\left[ 1+\frac{P_{i}\cdot h_{i,m}(t)}{\sigma^{2}}\right],
    \label{eq3}
\end{equation}
where $B_{i,m}(t)$ is the transmission bandwidth between $U_{i}$ and $AP_{m}$, and $P_{i}$ is the transmission power of $U_{i}$. $\sigma$ is the noise power. The task data size of $U_{i}$ is denoted as $X_{i}(t)$. Let $T_{i,m}^{up}(t)$ denote the communication delay of $U_{i}$ uploading its data to $AP_{m}$ and it can be expressed as:
\begin{equation}
   \hspace{4em}T_{i,m}^{up}(t)=\frac{X_{i}(t)}{R_{i,m}(t)}\cdot w_{im}(t).
   % \ (w_{nm}\in\left \{ 0,1 \right \} )
    \label{eq4}
\end{equation}

Let $I_{i,m}^{up}(t)$ denote the energy consumption of $U_{i}$ offloading task data to $AP_{m}$, which is formulated as:
\begin{equation}
    \hspace{3.5em}I_{i,m}^{up}(t)=X_{i}(t)\cdot \varsigma_{i,m}\cdot w_{im}(t), 
    \label{eq5}
\end{equation}
where $\varsigma_{i,m}$ is a constant that represents the unit energy consumption of $U_i$ uploading data to $AP_{m}$.

Second, we introduce the communication cost of graph data transmission between edge servers. When the edge server executes GNN inference, if the graph data of $SV_{k}$ is associated with that of $SV_{l}$, they need to communicate with each other to transmit related graph data. Let $R_{k,l}$ denote the transfer rate between $SV_{k}$ and $SV_{l}$, which can be given by:
\begin{equation}
   R_{k,l}=B_{k,l}\log_{2}\left[ 1+\frac{P_{k}\cdot h_{0}}{\sigma^{2}}\right], \forall k,l\in M, k\neq l, 
    \label{eq6}
\end{equation}
where $B_{k,l}$ is the transmission bandwidth between $SV_{k}$ and $SV_{l}$, and $P_{k}$ is the transmission power of $SV_{k}$. At time step $t$, we assume that $x_{k \rightarrow l}(t)$ is the data size that $SV_{k}$ transfers to $SV_{l}$, formulated as $\sum_{i=1}^{N} X_i \cdot w_{ik} \cdot e_{ij}$. Let $\tilde{x}_{kl}(t)$ denote the transmitted data between $SV_{k}$ and $SV_{l}$, where $\tilde{x}_{kl}(t) = x_{k \rightarrow l}(t) + x_{l \rightarrow  k}(t)$. Let $T_{k,l}^{tran}(t)$ denote the communication delay between $SV_{k}$ and $SV_{l}$ and it can be expressed as:
\begin{equation}
   \hspace{4.7em}T_{k,l}^{tran}(t)=\frac{\tilde{x}_{kl}(t)}{R_{k,l}}\cdot \eta_{kl}(t).
   % \ (\eta_{kl}\in\left \{ 0,1 \right \} )
    \label{eq7}
\end{equation}

Accordingly, the energy consumption in the communication process between $SV_{k}$ and $SV_{l}$ is given by $I_{k,l}^{com}(t)$, it can be represented as:
\begin{equation}
    % I_{k,l}^{com}(t)= \varsigma_{k,l}\cdot e_{ij}(t)\cdot w_{ik}(t)\cdot w_{jl}(t)\cdot \eta_{kl}(t) \cdot \tilde{x}_{kl}(t),
    \hspace{-1em}I_{k,l}^{com}(t)= \varsigma_{k,l}\cdot e_{ij}(t)\cdot w_{ik}(t)\cdot w_{jl}(t)\cdot \eta_{kl}(t) \cdot X_{i}(t),
    % I_{k,l}^{com}(t)=\varsigma_{k,l}\cdot e_{ij}(t)\cdot w_{ik}(t)\cdot w_{jl}(t)\cdot \eta_{kl}(t) \cdot X_{tr}(t),
    \label{eq8}
\end{equation}
where $\varsigma_{k,l}$ is a constant that represents the unit energy cost of data transmission between $SV_{k}$ and $SV_{l}$. 
% Eq.\ref{eq8} 
Eq.~\eqref{eq8}
indicates that When performing GNN inference, communication between $SV_{k}$ and $SV_{l}$ is required, because the associated $U_{i}$ and $U_{j}$ offload their tasks to $SV_{k}$ and $SV_{l}$ respectively.

\subsection{GNN computation cost}
In this subsection, we first consider the GNN computation time on the edge servers. Let $f_{k}$ denote the CPU clock cycles per second used by GNN on $SV_{k}$ to process the unit data. And on $SV_{k}$, the time required for the task of $U_{i}$ during GNN inference is denoted by $T_{i,f_{k}}^{com}(t)$, which can be expressed as:
\begin{equation}
    \hspace{4.7em}T_{i,f_{k}}^{com}(t)=\frac{X_{i}(t)\cdot w_{ik}}{f_{k}}.
    \label{eq9}
\end{equation}

Second, we consider the energy consumption of GNN computation. GNN are usually multi-layer and the computation process contains aggregation and update phases. Let $F$ be the number of layers in the GNN, and let $\kappa$ denote the $\kappa$-th layer of the GNN. At time step $t$, the energy consumption $I_{agg_{\kappa}}(t)$ of the aggregation phase in the $\kappa$-th layer of the GNN can be expressed as:
\begin{equation}
    \hspace{2.6em}I_{agg_{\kappa}}(t)=\sum_{i=1}^{N'}\mu\left|\mathcal{N}_{i}(t)\right|\cdot S_{\kappa-1}(t),
    \label{eq10}
\end{equation}
where $\mu$ is the unit data aggregation energy cost in GNN. $N'$ is the number of users in the current GNN computation. $|\mathcal{N}_{i}(t)|$ is the number of neighbors of $U_{i}$ at time step $t$, $S_{\kappa}(t)$ is the size of the feature data in the $\kappa$-th layer of the GNN, and the size of the original feature data is given by $S_{0}(t)$.

Let $I_{upd_{\kappa}}(t)$ denote the energy cost for the update phase of the $\kappa$-th layer of the GNN, it can be expressed as:
\begin{equation}
    \hspace{2em}I_{upd_{\kappa}}(t)=\vartheta S_{\kappa -1}(t) \cdot S_{\kappa}(t) + \varphi S_{\kappa}(t),
    \label{eq11}
\end{equation}
where $\vartheta$ is the cost of unit data for the GNN update phase, and $\varphi$ is the cost of unit data for multiplication when applying the activation function.

\subsection{Formulation of the problem}

Our goal is to obtain the optimal graph offloading strategy to minimize
the time and energy cost for the EC system to complete all user tasks. Let $T_{all}$ denote the total time required for the system to complete all user tasks. It includes the time needed for graph data offloading, graph data processing, and graph data transmission between edge servers. $T_{all}$ can be expressed as:
\begin{equation}
    T_{all}=\sum_{n=1}^{N}(\sum_{m=1}^{M}T_{n,m}^{up}+\sum_{k=1}^{M}\sum_{l=1,l\ne k}^{M}T_{k,l}^{tran}+\sum_{k=1}^{M}T_{n,f_{k}}^{com}).
    \label{eq12}
\end{equation}

Let $I_{all}$ denote the total energy consumption required for the EC system to complete all user tasks, which includes the offloading, aggregation, and update energy consumption of graph data of all GNN layers on the edge servers. $I_{all}$ can be expressed as:
\begin{equation}
    I_{all}=\sum_{n=1}^{N}( \sum_{m=1}^{M}I_{i,m}^{up}+\sum_{k=1}^{M}\sum_{l=1,l\ne k}^{M}I_{k,l}^{com})+\sum_{\kappa=1}^{F}(I_{agg_{k}}+I_{upd_{\kappa}}).
     % I_{all}=\sum_{n=1}^{N}\left ( \sum_{m=1}^{M}I_{i,m}^{up}+\sum_{k=1}^{M}\sum_{l=1,l\ne k}^{M}I_{k,l}^{com} \right ) +\sum_{\kappa=1}^{F}\left ( I_{agg_{k}}+I_{upd_{\kappa}} \right ) 
    \label{eq13}
\end{equation}

Let $\mathcal{C}$ represent the total cost of the system to accomplish all user tasks, i.e., $\mathcal{C}=T_{all}+I_{all}$. Based on this, given the edge network $\omega$ and the user's graph layout $\mathcal{G}$ perceived by the EC controller, our optimization objective is to find the optimal task offloading strategy $w$ for the user's task to minimize the task processing time and the energy consumption of the EC system, and the corresponding objective function $\mathcal{P}$ can be expressed as:
\vspace{-0.5em}
\begin{equation}
    \hspace{3.5em}\mathcal{P}=\sum_{t=1}^{\mathcal{T}} \underset{w}{\min}\ \mathcal{C}\left (w|\omega,\mathcal{G},t\right ). 
    \label{eq14}
\end{equation}
\vspace{-2em}
\begin{subequations}
    \begin{align}
    s.t. \hspace{2em}
        \label{eq14a}
        % & C1:e_{ij}\in\left \{ 0,1 \right \},\forall i,j\in N,i\ne j \tag{14a}\\
        % \label{eq14b}
        % & C2:\eta_{kl}\in\left \{ 0,1 \right \},\forall k,l\in M,k\ne l  \tag{14b}\\
        % \label{eq14c}
        & C1:\sum_{k=1}^{M} w_{ik}=1, \forall i\in N \tag{14a} \\
        \label{eq14d}
        & C2:f_{k}\ge 0,\forall k\in M \tag{14b}\\
        \label{eq14e}
        & C3:\sum_{i=1}^{N}\sum_{m=1}^{M}B_{i,m}\le B_{max1} \tag{14c} \\
        \label{eq14f}
        & C4:\sum_{k=1}^{M}\sum_{l=1}^{M}B_{k,l}\le B_{max2}, k\ne l \tag{14d} \\
        \label{eq14g}
        & C5:\sum_{i = 1}^{N}P_{i}\le P_{max1} \tag{14e} \\
        \label{eq14h}
        & C6:\sum_{k = 1}^{M}P_{k}\le P_{max2} \tag{14f},  
    \end{align}
\end{subequations}
where Constraint C1 means that each user's task can only be offloaded to one edge server. Constraint C2 indicates that the server clock frequency used by the GNN to process unit data must be greater than 0. Constraints C3 and C4 mean that the bandwidth between the user and the AP, as well as between edge servers, cannot exceed the maximum bandwidth allocated by the EC system. Constraints C5 and C6 indicate that the transmission power of both the user and the edge server cannot exceed the maximum transmission power allocated by the EC system.

Since the variable $w =\left \{ w_{ik}|1\le i \le N, 1\le j \le M \right \} $ in Eq.~\eqref{eq14} is the set of integer offloading decision variables that we require to solve, it contains a set of integer variables. And $w_{ik}$ has been explained in Section \ref{sec:sys-overview}. So the proposed optimization problem is an integer programming problem that is difficult to solve by conventional methods, we decouple it into two subproblems. (1) Optimize the original graph layout to  ensure that cross-server communication cost is minimized during GNN inference. Specifically, vertices corresponding to users with strong associations are cut into the same subgraph. Let $\mathcal{P}_{1}$ represent the graph layout optimization problem. (2) Based on the optimized graph layout, explore the graph offloading strategy to ensure that the total task processing time and energy cost of the EC system for processing all user tasks are minimized. Let $\mathcal{P}_{2}$ denote the graph offloading strategy exploration problem. And $\mathcal{P}_{1}$ can be denoted as:
\begin{equation}
\begin{array}{l}
\mathcal{P}_{1}=\sum_{t=1}^{\mathcal{T}}\sum_{k=1}^{M}\sum_{l=1,l\ne k}^{m} \underset{\mathcal{G}_{sub}}{min}\ I_{k,l}^{com}(\mathcal{G}_{sub}|\omega,G,t), \\
\hspace{1.3em}s.t. \hspace{2em} C1,
\end{array}
\tag{15}
\label{eq15}
\end{equation}
where $\mathcal{G}_{sub}$ is the optimized graph layout, which consists of a series of weakly associated subgraphs. The subproblem $\mathcal{P}_{2}$ can be expressed as:
\begin{equation}
\begin{array}{l}
\hspace{3em}\mathcal{P}_{2}=\sum_{t=1}^{\mathcal{T}}\underset{w}{min}\ C(w|\mathcal{G}_{sub},t), \\
\hspace{0em}s.t. \hspace{2em} C2-C6.
\end{array}
\tag{16}
\label{eq16}
\end{equation}

$\mathcal{P}_{1}$ and $\mathcal{P}_{2}$ will be detailed in Section~\ref{HiCut} and Section~\ref{DRLGO}.

\section{Hierarchical traversal graph cut algorithm}
\label{HiCut}
In this section, we will describe the hierarchical traversal graph cut algorithm, HiCut, Which is used to solve the graph layout optimization problem $(\mathcal{P}_{1})$. Firstly, we introduce the graph cut insight of the algorithm. Then the process of the algorithm will be introduced. Finally, we analyze the complexity of the algorithm.
\subsection{Graph cut insights}
\label{HiCut:insights}
According to Section~\ref{sec:sys-overview}, the topology formed by users in the EC system is perceived as graph layout $\mathcal{G}$. In this graph layout, let $|\mathcal{N}_{i}|$ be the number of users that $U_{i}$ is associated with.
When the task data of $U_{i}$ is in GNN inference, it needs to obtain task data from the $|\mathcal{N}_{i}|$ associated users, and if these task data from associated users are offloaded to other edge servers, the edge server $U_{i}$ located will communicate with edge servers where associated users are located for transferring task data. And this will result in a large amount of data communication cost between edge servers.

It can be seen that the above problem is caused by the fact that in the graph layout $\mathcal{G}$, the task data of $U_{i}$ and its associated users are not offloaded to the same edge server. However, the deeper reason is that GNN inference requires aggregating the task data for a user and its associated users. Therefore, we consider optimizing the graph layout $\mathcal{G}$ according to the aggregation characteristics of GNN. To this end, we propose the hierarchical traversal graph cut algorithm, HiCut, which adjusts the graph layout $\mathcal{G}$ to a new graph layout. Let $\mathcal{G}_{sub}$ represent the new graph layout and $\mathcal{G}_{sub_{c}}$ represent a subgraph in $\mathcal{G}_{sub}$. The new graph layout $\mathcal{G}_{sub}$ consists of a series of subgraphs, where vertices in a subgraph are strongly associated, while vertices between different subgraphs are weekly associated. If a user with more associations with other users, it indicates a stronger association with those users. When exploring graph offloading strategies based on $\mathcal{G}_{sub}$, since the data associations between different subgraphs are minimized, the cross-server communication cost between the edge servers is also minimized. The $\mathcal{G}_{sub}$ and $\mathcal{G}_{sub}$ can be expressed as:
\begin{equation}
\begin{array}{l}
\hspace{2.5em} \mathcal{G}_{sub}=\left \{\mathcal{G}_{sub_{c}}|c=1,2,...,C\right \},
\end{array}
\tag{17}
\label{eq17}
\end{equation}
% and
% \vspace{-2em}
\vspace{-0.5em}
\begin{equation}
\begin{array}{l}
 \hspace{1.2em}\mathcal{G}_{sub_{c}}=\left \{V_{i},\dots,V_{j}|\forall i,j\in N,i\ne j\right \},
\end{array}
\tag{18}
\label{eq18}
\end{equation}
\noindent where $C$ is the number of subgraphs in $\mathcal{G}_{sub_{c}}$.

\begin{figure}[htbp]
    \centering
    \includegraphics[width=8cm,height=4.5cm]{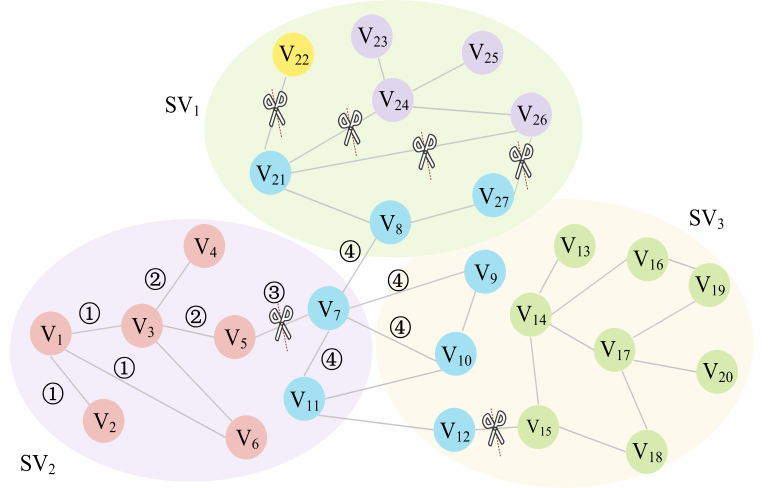}
    \caption{The Cut Process of HiCut.}
    \label{block}
\end{figure}
\vspace{-1em}
\subsection{The process of HiCut}
First of all, HiCut is based on the graph's breadth-first search (BFS) algorithm~\cite{bader2006designing} and layer-by-layer traversal graph layout $\mathcal{G}$. Although the graph's depth-first search (DFS) algorithm~\cite{riansanti2018connectivity} can also traverse the graph, it is limited by stack memory and difficult to assess the associations between the current vertex and vertices outside the recursion path, so we choose the BFS as the basis of HiCut.

Secondly, the graph cut operation should be performed between the two layers with the weakest vertex associations. The associations between vertices in different layers are represented by the number of edges. Specifically, when finding the position of the graph cut operation, starting from the second layer, we compare the number of edges in the current layer, $d_{n}$, with that in the previous layer, $d_{n-1}$, n is the number of the traversal layer. If the number of edges increases, it indicates that the associations between these two layers is strengthening, so the vertices in both layers should be placed in the same subgraph, and the traversal continues. If the number of edges decreases, we record the vertices of the current layer in the variable $V_{seg}$ and continue traversing until the number of edges between two layers increases again. At this point, the most recently recorded $V_{seg}$ marks the optimal graph cut position, ensuring minimal inter-subgraph edges after the graph cut operation. Since the associations between vertices in the layers after the graph cut position begins to strengthen again, the graph cut position found based on the above criteria ensures that the correlation between subgraphs is minimized.

Thirdly, After finding the graph cut position and performing the graph cut operation, we obtain a subgraph in $\mathcal{G}_{sub}$. By using the vertices not in $G_{sub}$ as starting vertex and repeatedly performing the above graph cut operations, the optimized graph layout $G_{sub}$ can be obtained.

Finally, Fig.~\ref{block} shows the graph layout optimization process of HiCut, we use the same color to represent the vertices within a subgraph, and it can be seen that the original graph layout $\mathcal{G}$ is finally cut into 5 subgraphs. Next, we take the subgraph with red vertices as an example to explain how HiCut performs the graph cut operation on $\mathcal{G}$. Let $\mathcal{G}_{sub_{c}}$ denote the subgraph with red vertices. In Fig.~\ref{block}, the numbers on the edges represent the traversal layer's number. At the beginning of the graph cut operation, $V_{1}$ is first traversed by HiCut as the starting vertex, at which $d_{1}$ is 3, since no comparison is performed in the first layer, $V_{1}$ is directly added to $\mathcal{G}_{sub_{c}}$ and then the traversal proceeds to the second layer. In the second layer, the value of $d_{2}$ is 2, it satisfies the condition $d_{n} < d_{n-1}$, so the second layer can be considered as a graph cut position. However, to ensure minimal association between subgraphs, we save the current layer's vertices into $V_{seg}$ and continue the traversal. In the third layer, $d_{n}$ is 1, which satisfies the condition $d_{n} < d_{n-1}$. Due to the fact that there are vertices in $V_{seg}$, we first save the vertices in $V_{seg}$ (i.e., $V_{2}$, $V_{3}$, and $V_{6}$) into $\mathcal{G}_{sub_{c}}$, and the remaining steps are the same as in the second layer. Finally, in the fourth layer, $d_{n}=4$ satisfies the condition $d_{n} \geq d_{n-1}$, and $V_{seg}$ contains vertices. Therefore, we add the vertices in $V_{seg}$ (i.e., $V_{4}$ and $V_{5}$) to $\mathcal{G}_{sub_{c}}$ and end the current graph cut operation. Thus, we obtain the subgraph $\mathcal{G}_{sub_{c}} = \{V_{1}, V_{2}, V_{3}, V_{4}, V_{5}, V_{6}\}$ composed of the red vertices. Finally, we can obtain all the subgraphs shown in Fig.~\ref{block} by repeating the graph cut operation with the vertices that are not in $\mathcal{G}_{sub}$ as the starting vertices.

\subsection{Description of HiCut}
The detailed procedures of HiCut are summarized in Algorithm~\ref{alg1}. HiCut accepts the original graph layout $\mathcal{G}$ and $N$ vertices as input, these vertices represent users, which are denoted as $V$. The output of HiCut is the optimized graph layout $\mathcal{G}_{sub}$. Additionally, the execution of the HiCut requires several extra variables. Specifically, $\mathcal{Q}$ is a queue used to store the vertices of each layer in $\mathcal{G}$ and perform the layer-by-layer traversal of $\mathcal{G}$. $\mathcal{G}_{sub_{c}}$ is used to represent the subgraph obtained for the current graph cut operation. $N_{cur}$ denotes the number of vertices in the current layer, and its initial value is 1. $l_{cur}$ is the current layer's number, and its initial value is 1. $V_{cur}$ is the set of vertices in the current layer. $V_{seg}$ is a variable for recording vertices of graph cut position. The initial values of $d_{n-1}$ and $d_{n}$ are all 0. \texttt{LayerCut($\cdot$)} is the main graph cut function of Algorithm~\ref{alg1}, and it terminates when $\mathcal{G}$ is empty or the garph cut operation for get $\mathcal{G}_{sub_{c}}$ is completed. HiCut ends when $\mathcal{G}$ is empty or $\mathcal{G}$ is fully processed. Since the graph cut process has been detailed, we will now introduce the main execution steps of the algorithm. First, HiCut traverses all vertices in $\mathcal{G}$ and calls the function \texttt{LayerCut($\cdot$)} with vertices not in $\mathcal{G}_{sub}$ as the starting points (lines 1-4). Next, the function \texttt{LayerCut($\cdot$)} mainly consists of two parts: hierarchical traversal of $\mathcal{G}$ and execute the graph cut operation. In the function \texttt{LayerCut($\cdot$)}, if the exit condition is not satisfied, first initialize the needed variables, and perform some initial operations (lines 8-10). Next, the BFS algorithm is used to perform a layer-by-layer traversal of $\mathcal{G}$ with a queue $\mathcal{Q}$ (lines 11-19). And then, the optimal graph cut position is iteratively found during the traversal process, until complete the current graph cut operation(lines 20-36). Finally, after the processing of HiCut, we can get the optimized graph layout $\mathcal{G}_{sub}$.

\begin{algorithm}[t]
% \setlength{\algomargin}{15pt} 
% \LinesNotNumbered % 特定算法中不显示行号
\DontPrintSemicolon
\caption{Hierarchical Traversal Graph Cut Algorithm}
\label{alg1}
\KwIn{
    $\mathcal{G}$ : The graph layout of users \;
    \hspace{3.3em}$\mathcal{V}$ : The set of vertices in the graph layout
}
\KwOut{$\mathcal{G}_{sub}$ : the set of subgraphs}
\SetKwFunction{LayerCut}{LayerCut} 
Initialize the empty set of subgraphs $\mathcal{G}_{sub}$ \;
\hspace*{-0.35em} \For{$\mathcal{V}_{i}$ in $\mathcal{V}$}{
    \If{$\mathcal{V}_{i}$ not in $\mathcal{G}_{sub}$}{
        \LayerCut($\mathcal{G}$, $\mathcal{V}_{i}$, $\mathcal{V}$, $\mathcal{G}_{sub}$)\;
    }
}

\SetKwProg{Fn}{Function}{:}{}
\Fn{\LayerCut{$\mathcal{G}$, $\mathcal{V}_{begin}$, $\mathcal{V}$, $\mathcal{G}_{sub}$}}{
    \If{$\mathcal{G}$ is empty}{
        exit\;
    }
    Initialize variables: $\mathcal{G}_{sub_{c}}$, $\mathcal{Q}$, $N_{cur} = 1$, $l_{cur} = 1$, $V_{cur}$, $V_{seg}$,  $d_{n-1}$ = $d_{n}$ = 0 \;
    Add $\mathcal{V}_{begin}$ to $\mathcal{G}_{sub_{c}}$, $\mathcal{Q}$ \;
    Add $\mathcal{G}_{sub_{c}}$ to $\mathcal{G}_{sub}$ \;
    \While{$\mathcal{Q}$}{
        current vertex $\mathcal{V}_{c} \gets \texttt{front}(\mathcal{Q})$ \;
        Add $\mathcal{V}_{c}$ to $V_{cur}$ \;
        $N_{cur} \gets N_{cur} - 1$, \texttt{removeFront}($\mathcal{Q}$) \;
        \For{$\mathcal{V}_{r}$ in neighbors of $ \mathcal{V}_{c}$} {
            \If{$\mathcal{V}_{r}$ not in $\mathcal{G}_{sub}$}{
                $d_{n} \gets d_{n} + 1$ \;
            }
            \If{$\mathcal{V}_{r}$ not visited}{
                \texttt{enqueue}($\mathcal{Q}$, $\mathcal{V}_{r}$)\;
            } 
        }
        \If{$N_{cur}$ == 0}{
            $N_{cur} \gets \texttt{length}(\mathcal{Q})$\;
            \If{$d_{n} == 0$}{ 
                Add $V_{seg}, V_{cur}$ to $\mathcal{G}_{sub_{c}}$ and exit \;
            }
            \eIf{$l_{cur}$ == 1}{
                $d_{n-1} \gets d_{n}$\;
            }{
                \If{$d_{n-1} <= d_{n}$}{
                    \eIf{$V_{seg}$ and $d_{n-1} < d_{n}$}{
                        Add $V_{seg}$ to $\mathcal{G}_{sub_{c}}$ and exit \;
                    }
                    {
                        $d_{n-1} \gets d_{n}$, add $V_{cur}$ to $\mathcal{G}_{sub_{c}}$ \;
                    }
                }
                \If{$d_{n-1} > d_{n}$}{
                    \If{$V_{seg}$}{
                        Add $V_{seg}$ to $\mathcal{G}_{sub_{c}}$
                    }
                    \texttt{clear}($V_{seg}$), $V_{seg} \gets V_{cur}$\;
                    $d_{n-1} \gets d_{n}$
                }
            }
        }
        $l_{cur} \gets l_{cur} + 1$, \texttt{clear}($V_{cur}$), $d_{n} \gets 0$\;
    }
}
\end{algorithm}

\subsection{Complexity analysis}
The complexity analysis of algorithm~\ref{alg1} consists of two aspects: (1) The traversal of user corresponding vertices (lines 2-4). (2) The invocation of \texttt{LayerCut($\cdot$)} (lines 5-36). First, for the traversal of vertices, we need to check whether the current vertex $V_{i}$ is in $\mathcal{G}_{sub}$. Since the vertices in $\mathcal{G}$ correspond one-to-one with users, it is necessary to perform $N$ traversals and the computational complexity of this process is $O(N)$. Second, the function \texttt{LayerCut($\cdot$)} is based on the BFS algorithm for graphs. The complexity of the BFS algorithm is $O(N + E)$~\cite{bader2006designing}. Unlike BFS, the \texttt{LayerCut($\cdot$)} includes the step of determining the graph cut positions (lines 20-36). But the determination of the graph cut position are conditional statements, which can be completed in constant time. Thus, its computational complexity is $O(1)$. Due to the function \texttt{LayerCut($\cdot$)} is invoked during the vertex traversal process, the overall computational complexity of Algorithm~\ref{alg1} is \(O(N) \times O(N + E)\), i.e., \(O(N^2 + NE)\)
% which simplifies to \(O(N^2 + NE)\).

\section{DRL-based graph offloading algorithm}
\label{DRLGO}
In this section, we introduce the DRL-based graph offloading algorithm, DRLGO, which is designed to solve the second subproblem $(\mathcal{P}_{2})$. How to build the reinforcement learning environment for the training of agents and how to update the policy of agents will be detailed.

\subsection{Adaptive graph offloading problem}
The original graph layout $\mathcal{G}$ is perceived by the EC controller, after being optimized by HiCut, the new graph layout $\mathcal{G}_{sub}$ contains a series of subgraphs. Furthermore, during GNN inference, the message passing between these subgraphs is minimized. However, we still face the challenge of how to offload the user tasks corresponding to the vertices of a subgraph to the same edge server as much as possible, i.e., the graph offloading problem. At the same time, to minimize the cost of the EC system, the graph offloading strategy needs to fully consider complex information such as user location, communication bandwidth, service capabilities of edge servers, and dynamic states of users in the EC system. Undoubtedly, the exploration of graph offloading strategy is quite complex and it is difficult to use traditional methods to find the optimal strategy. Therefore,  we use the DRL method to solve it and use the reward function to constrain the vertices of the same subgraph to be offloaded to the same edge server as much as possible. Considering that in multi-agent reinforcement learning, centralized trained agents can be deployed in a distributed manner to run on different edge servers, it satisfies distributed GNN inference and collaborative communication between edge servers. Moreover, each edge server can be considered as an agent and cooperate with other agents, and then centrally trained to obtain the optimal graph offloading strategy to minimize the system cost, so we build DRLGO based on MADDPG~\cite{lowe2017multi}. Next, there will be a detailed description of how to build the environment of the multi-agent Markov decision process (MAMDP)~\cite{tsaousoglou2022multistage} 
and how the agent update its policy.

\subsection{MAMDP environment}
In DRLGO, agents need to continuously interact with the environment to learn the optimal graph offloading strategy. The environment model is represented by $(\mathcal{S},\mathcal{A},\mathbb{T},\mathcal{R},\mathcal{S}_{0})$. During the training process, all users are iterated one by one, and at each iteration, all agents make offloading actions at the same time, but the user task can only be offloaded to one edge server. To ensure that the user tasks corresponding to the vertices of a subgraph are offloaded to the same edge server, we applied constraints through the reward function. The following is a detailed definition of each part of the MAMDP environment.

\begin{description}[labelindent=2em, leftmargin=0pt, labelsep=1em]
    \item[a) \textit{Environment states}:] The state of the environment which is given by $\mathcal{S}(t)$ consists of local observations of all agents. Let $\mathcal{O}_{m}$ denotes the observation of agent $m$.  $\mathcal{S}(t)$ includes the location of each user, the number of neighbors of each user, the size of the user's task data, 
    % user's task offloading information which denoted as $b$ $(b \in \left \{ 0,1 \right \})$, 
    the bandwidth of the links from the users to the APs, the service capabilities of the edge servers and the bandwidth of links between edge servers. The $\mathcal{S}(t)$ and $\mathcal{O}_{m}$ can be expressed as: 
\end{description}
\begin{equation}
    \begin{aligned}
        % \mathcal{S}(t) = \{ & x_{1}(t), y_{1}(t), \ldots, x_{N}(t), y_{N}(t), |\mathcal{N}_{1}(t)|,\ldots,\\
        % &|\mathcal{N}_{N}(t)|,X_{1}(t), \ldots, X_{N}(t), b_{1}(t),\ldots, b_{N}(t)\\
        % &, C_{1}(t), \ldots, C_{M}(t),B_{11}(t),\dots,B_{im}(t),\dots,\\
        % &B_{kl}(t),\dots,B_{MM}|\forall i \in N,m,k,l \in M\},
        \mathcal{S}(t) = \{ & x_{1}(t), y_{1}(t), \ldots, x_{N}(t), y_{N}(t), |\mathcal{N}_{1}(t)|,\ldots,\\
        &|\mathcal{N}_{N}(t)|,X_{1}(t), \ldots, X_{N}(t), C_{1}(t), \ldots,\\
        &C_{M}(t),B_{11}(t),\dots,B_{im}(t),\dots,B_{kl}(t),\dots,\\
        &B_{MM}|\forall i \in N,m,k,l \in M\},
    \end{aligned}
    \tag{19}
    \label{eq19}
\end{equation}
\vspace{-1em}
\begin{equation}
    \begin{aligned}
       % \hspace{0em} \mathcal{O}_{m}(t) = \{ & x_{i}(t), y_{i}(t), \ldots, x_{j}(t), y_{j}(t), |\mathcal{N}_{i}(t)|,\ldots,\\
       % & |\mathcal{N}_{j}(t)|,X_{i}(t), \ldots, X_{j}(t),b_{i}(t),\ldots,b_{j}(t),\\
       % & C_{m}(t),B_{i1},\dots,B_{jM}|\forall i,j\in N\},
        \hspace{0em} \mathcal{O}_{m}(t) = \{ & x_{i}(t), y_{i}(t), \ldots, x_{j}(t), y_{j}(t), |\mathcal{N}_{i}(t)|,\ldots,\\
       & |\mathcal{N}_{j}(t)|,X_{i}(t), \ldots, X_{j}(t),C_{m}(t),B_{i1},\dots,\\
       &B_{jM}|\forall i,j\in N\},
    \end{aligned}
    \tag{20}
    \label{eq20}
\end{equation}
where we assume that users numbered $i$ to $j$ are in the service scope of the edge server where agent $m$ is located. In addition, each agent can only observe the information within the service scope of its located server.

\begin{description}[labelindent=2em, leftmargin=0pt, labelsep=1em]
    \item[b) \textit{Environment actions}:] 
    % The actions of the environment include the global action $A$ of all agents and the action $A_{m}$ of the agent m. At moment t, $A(t)$ and $A_{m}(t)$ can be expressed as: 
    Each agent interacts with the environment using local action and trains using global action. And the global action and local action are denoted as $\mathcal{A}$ and $\mathcal{A}_{m}$, where $m$ is the number of the agent. At time step t, $\mathcal{A}(t)$ and $\mathcal{A}_{m}(t)$ can be expressed as:
\end{description}
\begin{equation}
    \begin{aligned}
       \hspace{3.7em} \mathcal{A}(t)=\left \{ \mathcal{A}_{1}(t),\dots,\mathcal{A}_{Q}(t) \right \},
    \end{aligned}
    \tag{21}
    \label{eq21}
\end{equation}
\vspace{-1.5em}
\begin{equation}
    \begin{aligned}
      \mathcal{A}_{m}(t)=(\mathcal{A}_{m1}(t),\mathcal{A}_{m2}(t)) \ (\mathcal{A}_{m1},\mathcal{A}_{m2}\in [0,1]),
    \end{aligned}
    \tag{22}
    \label{eq22}
\end{equation}
where the action $A_{m}$ is two-dimensional, and the dimension with the maximum value in $A_{m}$ is taken as the offloading decision, which determines whether the task of the current user will be offloaded to the server where agent $m$ located.

\begin{description}[labelindent=2em, leftmargin=0pt, labelsep=1em]
    \item[c) \textit{Environment rewards}:] According to the EC system's cost function $C = T_{all} + I_{all}$, let  $C_{m}(t)$ represent the action reward of agent $m$ at time step $t$, it reflects the time and energy cost for processing the user task. Additionally, the negative sum of all agents' rewards is used as the global reward at time step $t$. Let $\mathcal{R}(t)$ and $\mathcal{R}_{m}(t)$ represent the global reward and the reward for agent $m$ at time step $t$, respectively. Both of them can be expressed as:
\end{description}
\begin{equation}
    \begin{aligned}
      \hspace{5em}\mathcal{R}(t)=\sum_{m=1}^{M}\mathcal{R}_{m}(t),
    \end{aligned}
    \tag{23}
    \label{eq23}
\end{equation}
\vspace{-1em}
\begin{equation}
    \begin{aligned}
      \hspace{3.3em}\mathcal{R}_{m}(t)=-(C_{m}(t)+\mathcal{R}_{sp}(t)),
    \end{aligned}
    \tag{24}
    \label{eq24}
\end{equation}
where $\mathcal{R}_{sp}(t)$ denotes whether the tasks of users corresponding to the vertices within a subgraph are offloaded to the same edge server as much as possible. And the more edge servers the tasks are offloaded to, the larger the value of $\mathcal{R}_{sp}(t)$. For the subgraph $\mathcal{G}_{sub_{c}}$ with index $c$, let $\mathbb{N}_{c}$ represent the number of users in $\mathcal{G}_{sub_c}$ whose tasks have been offloaded, and $\mathbb{N}_{s}$ represent the number of edge servers to which the tasks in $\mathcal{G}_{sub_c}$ have been offloaded. $\mathcal{R}_{sp}(t)$ can be represented as:
\begin{equation}
    \begin{aligned}
      \hspace{5em}\mathcal{R}_{sp}(t)= \zeta \cdot \frac{\mathbb{N}_{s}(t)}{\mathbb{N}_{c}(t)},
    \end{aligned}
    \tag{25}
    \label{eq25}
\end{equation}
where $\zeta$ is a weight constant.

\begin{description}[labelindent=2em, leftmargin=0pt, labelsep=1em]
    \item[d) \textit{Transition probability}:]$\mathbb{T}$ represents the probability that the state transitions from $\mathcal{S}(t)$ to the next state $\mathcal{S'}(t)$ after agents choose the global action $\mathcal{A}(t)$, it can be denoted as: $\mathbb{T}=\left \{ p(\mathcal{S'}(t)|\mathcal{S}(t),\mathcal{A}(t))\  \forall \mathcal{S}(t),\mathcal{S'}(t) \in \mathcal{S},\mathcal{A}(t)\in\mathcal{A} \right \}$
\end{description}

\begin{description}[labelindent=2em, leftmargin=0pt, labelsep=1em]
    \item[e) \textit{Initial states}:] Let $\mathcal{S}_{0}$ denote the initial state of the environment, in which the tasks of all users are not offloaded. All edge servers are in the initial state. 
\end{description}

\subsection{MADDPG-based DRLGO}
MADDPG is an extension of DDPG \cite{lin2022data} and works well for continuous state and action spaces. Since the action output and the state space values are continuous, MADDPG is perfectly suitable for training in the environment we have built. In DRLGO, all agents cooperate to explore the optimal graph offloading strategy, and each agent uses a centralized training distributed execution framework. The critic network of each agent in the training stage can get the state and action info of other agents. However agents can only choose their actions based on their local observation, and the environment will give the reward for the actions of the agents. Once the global reward $\mathcal{R}$ converges, it indicates that each agent gets a trained policy and can be deployed for distributed execution~\cite{he2023fairness}. Furthermore, the transition probability $\mathbb{T}$ can be re-expressed as: 
\begin{equation}
    \begin{aligned}
        \mathbb{T} &= p(\mathcal{S'}(t) \mid \mathcal{S}(t), \mathcal{A}(t), \pi(t)) = p(\mathcal{S'}(t) \mid \mathcal{S}(t), \mathcal{A}(t)) \\
        &= p(\mathcal{S'}(t) \mid \mathcal{S}(t), \mathcal{A}(t), \pi'(t)),
    \end{aligned}
    \tag{26}
    \label{eq26}
\end{equation}
where $\pi(t)=\left \{ \pi_{1}(t),\dots,\pi_{M}(t) \right \}$ is the actor network for all agents, and $\pi'(t)=\left \{ \pi_{1}'(t),\dots,\pi_{M}'(t) \right \} $ is the target actor network for all agents, and $\theta' = \left \{ \theta_{1}',\dots,\theta_{M}' \right \}$ be the parameters of the target actor network $\pi'$. Meanwhile, let $\theta = \left \{ \theta_{1},\dots,\theta_{M} \right \} $ be the parameters of the actor network $\pi$. The cumulative expected reward for agent $m$ can be expressed as: 
\begin{equation}
    \begin{aligned}
   \hspace{3em}J(\theta_{m})=\mathbb{E}_{\mathcal{S},\mathcal{A}\sim D} \left [ \sum_{t=1}^{\mathcal{T}} \gamma \mathcal{R}_{m}(t)  \right ],
    \end{aligned}
    \tag{27}
    \label{eq27}
\end{equation}
where $D$ is the experience replay buffer which can be expressed as $\left \{ \mathcal{S},\mathcal{A},\mathcal{R},\mathcal{S}',done\right \}$. $\mathcal{R}$ stores the rewards of all agents, i.e. $\left \{ \mathcal{R}_{1},\dots,\mathcal{R}_{Q}\right \} $. $done=\{done_{1}, \dots, done_{m}\}$ is a set of boolean variables representing the exploration status of each agent. A value of $true$ indicates an active state, while $false$ indicates an inactive state. When all user tasks are offloaded, all values in $done$ are set to $true$. Additionally, when the edge server that the agent $m$ located reaches its maximum service capacity, $done_{m}$ is also set to $true$. $\gamma$ is the discount factor. The policy gradient of $\pi_{m}$ for the agent m is:
\begin{equation}
    \begin{aligned}
        \bigtriangledown_{\theta_{m}}J(\theta_{m}) &= \mathbb{E}_{\mathcal{S,A} \sim D} \left[ 
        \bigtriangledown_{\theta_{m}} \pi_{m}(\mathcal{A}_{m} \mid \mathcal{O}_{m}) 
        \bigtriangledown_{\mathcal{A}_{m}} \right. \\
        &\quad \left. \mathbb{Q}_{m}(\mathcal{S}, \mathcal{A}) \mid \mathcal{A}_{m} = \pi_{m}(\mathcal{O}_{m}) 
        \right],
    \end{aligned}
    \tag{28}
    \label{eq28}
\end{equation}

\begin{algorithm}[t]
\DontPrintSemicolon
\caption{The Training Process of DRLGO}
\label{alg2}

Initialize the actor network $\pi$ and target actor network $\pi'$ with weight $\theta$ and $\theta'$\;
Initialize the critic network $\mathbb{Q}$ and target critic network $\mathbb{Q}'$ with weight $\varphi$ and $\varphi'$\;
Initialize the size of replay buffer\;

\For{episode $e=1,\dots,E$}{
    Receive initial state $\mathcal{S}_{0}$\;
    \For{$t=1,2,\dots,\mathcal{T}$}{
        All elements in $done$ are initialized to \textbf{False}\;
        Dynamic change the environment and
        execute Algorithm \ref{alg1} to get $\mathcal{G}_{sub}$\;
        \While{not done}{
             For each agent $m$, select action $\mathcal{A}_{m}=\pi_{m}(\mathcal{O}_{m})$\;
             Execute global action $\mathcal{A}=\left \{ \mathcal{A}_{1},\dots,\mathcal{A}_{Q}\right \}$\;
             Receive reward $\mathcal{R}$ and next state $\mathcal{S'}$\;
             $done_{i} \gets \textbf{True}$ if server $i$ reach maximum service capacity\;
             Store ($\mathcal{S}$, $\mathcal{A}$, $\mathcal{R}$, $\mathcal{S'}$, $done$) to replay buffer\;
             \For{agent $m = 1$ to $Q$}{
                 Sample a mini batch of experience $\mathcal{B}$ as $ (\mathcal{S}, \mathcal{A}, \mathcal{R}, \mathcal{S'}, done)$ from replay buffer\;
                 Calculate the value of $\mathbb{Y}$ through Eq.~\eqref{eq30}\;
                 % Set the $\mathbb{Y}$ value as Eq.~\ref{eq30}\;
                 Update critic network by minimizing the loss: $\frac{1}{\mathcal{B}} \sum_{j=1}^{\mathcal{B}}\left [ (\mathbb{Q}_{m}(\mathcal{S}^{j},\mathcal{A}^{j})-\mathbb{Y}^{j})^2 \right ]$\;
                 % \text{Eq.~\ref{eq31}}$\;
                 Update the actor network policy using: $\frac{1}{\mathcal{B}} \sum_{j=1}^{\mathcal{B}} \bigtriangledown_{\theta_{m}}J(\theta_{m})$\;
             }
             Soft update parameters of each agent by:
             $ \theta_{m}'\leftarrow \tau \theta_{m}+(1-\tau)\theta_{m}'$
             $\varphi_{m}'\leftarrow \tau \varphi_{m}+(1-\tau)\varphi_{m}'$
        }
    }
}

\end{algorithm}

\noindent where $\varphi=\left \{ \varphi_{1},\dots,\varphi_{Q} \right \}$ is the parameter of the critic network $\mathbb{Q} $, and $\varphi'=\left \{ \varphi_{1}' ,\dots,\varphi_{Q}' \right \}$ is the parameter of target critic network  $\mathbb{Q}'$.
$\mathbb{Q}_{m}$ evaluates the expected value of the future cumulative rewards of the agent $m$ in centralized training. The mean square error is used as the loss function to evaluate the policy of actor network. The loss function of the agent $m$ can be denoted as:
\begin{equation}
    \begin{aligned}
    \hspace{1.5em}L(\varphi_{m})=\mathbb{E}_{\mathcal{S},\mathcal{A},\mathcal{R},\mathcal{S'}}\left [ (\mathbb{Q}_{m}(\mathcal{S},\mathcal{A})-\mathbb{Y})^2 \right ], 
    \end{aligned}
    \tag{29}
    \label{eq29}
\end{equation}
where $\mathbb{Y}$ is the target evaluation value, which can be expressed as:
\begin{equation}
    \begin{aligned}
    \hspace{0.7em}\mathbb{Y}=\displaystyle\sum_{t=1}^{\mathcal{T}}\mathcal{R}_{m}(t)+(1-done)\cdotp
    \gamma \mathbb{Q} _{m}'(\mathcal{S}',\mathcal{A}' ),
    \end{aligned}
    \tag{30}
    \label{eq30}
\end{equation}
where $\mathcal{A}'$ is the set of output actions of the target actor network, which can be represented as $\left \{ \pi_{1}'(\mathcal{O}_{1}'),\dots,\pi_{Q}'(\mathcal{O}_{Q}') \right \}$. $\mathbb{Q} _{m}'$ is the target critic network of the agent m. The parameters $\theta_{m}'$ and $\varphi_{m}'$ use the soft update with the following formulas:
\begin{equation}
    \begin{aligned}
    \hspace{4.4em}\theta_{m}'\leftarrow \tau \theta_{m}+(1-\tau)\theta_{m}',
    \end{aligned}
    \tag{31}
    \label{eq31}
\end{equation}
\vspace{-1.7em}
\begin{equation}
    \begin{aligned}
    \hspace{4.2em}\varphi_{m}'\leftarrow \tau \varphi_{m}+(1-\tau)\varphi_{m}',
    \end{aligned}
    \tag{32}
    \label{eq32}
\end{equation}
where $\tau$ is the soft update factor.

\begin{figure*}[htbp]
    \centering
    \includegraphics[width=\linewidth,height=7.5cm]{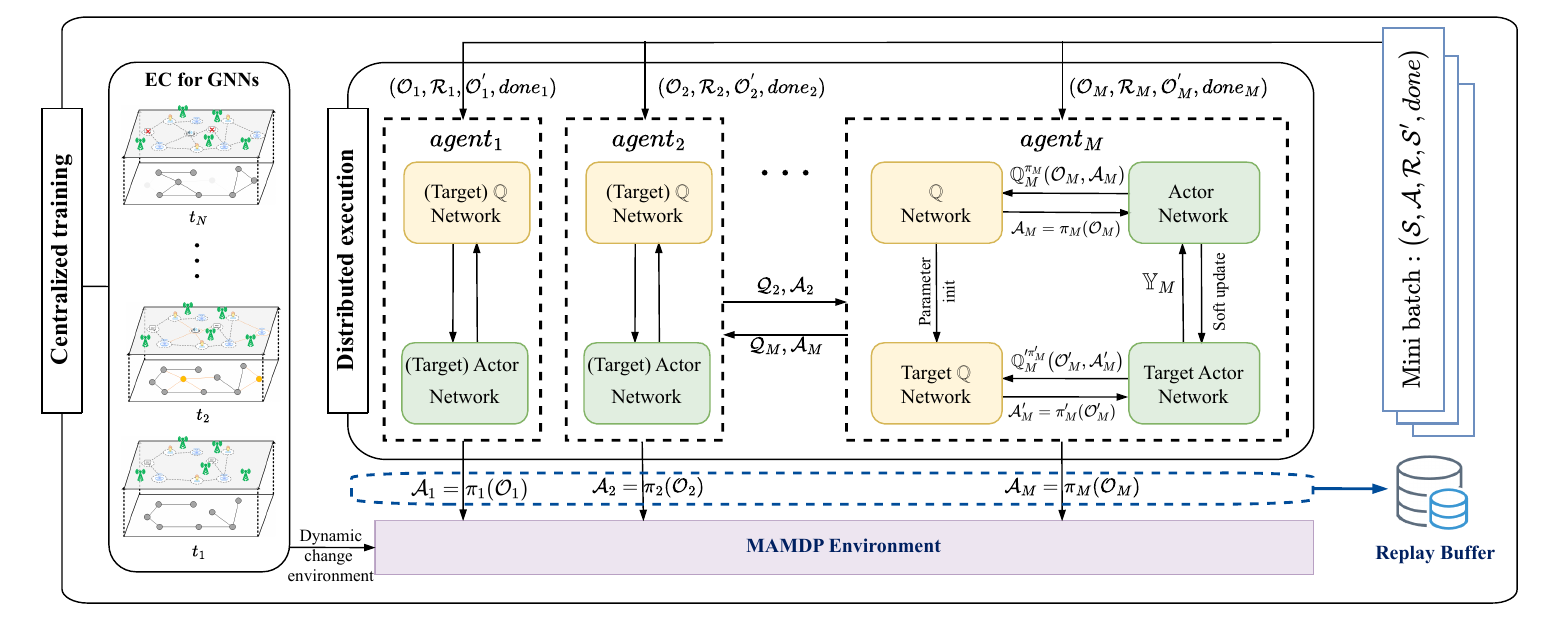}
    \caption{The training process of DRLGO.}
    \label{train}
\end{figure*}

As shown in Fig.~\ref{train}, multiple agents interact with the environment, and each agent with two main networks and two target networks, which are trained centrally. In each training episode, we randomly change the environment dynamically from the choices of increasing or decreasing the users, changing the associations of users, and changing the position of the users. Then our proposed dynamic graph model is constructed to represent the original graph layout and optimized by HiCut. Finally, train the DRLGO to explore the optimal graph offloading strategy. Algorithm~\ref{alg2} presents the detailed training process of DRLGO.

\section{Experiment}
\label{Experiment}
In this section, we evaluate and analyze the proposed method through multiple experiments. First, we compare the graph cut performance of proposed HiCut and \textit{minimum cut algorithm} in work \cite{zeng2022gnn}. This is a recent work and the maximum flow minimum cut algorithm is a classical and well-established graph cut algorithm that provides strong theoretical guarantees for graph cut. Then, under dynamic user states, the performance of the proposed DRLGO is compared with other task offloading methods. Next, we conducted a comparative analysis of the convergence of DRLGO and another method. Finally, the ablation experiment is conducted to validate the effectiveness of the graph offloading strategy of DRLGO.

% \begin{table}[width=\linewidth,cols=2,pos=h]
% \footnotesize
% \caption{Parameters of the system.}\label{table2}
% \begin{tabular*}{\tblwidth}{@{} CC@{} }
% \toprule
% Parameters & Value\\
% \midrule
% Noise power, $\sigma^2$ & -110 dBm \\
% Number of users, $N$ & [50,300] \\
% Transmission power of user $i$, $P_{i}$ & [2,5] mW \\
% Transmission power of edge server $k$, $P_{k}$ & [10,15] mW \\
% Unit data aggregation cost of GNN inference & 20 pJ/bit \\
% Unit data update cost of GNN inferenc, $\vartheta$ & 100 pJ/bit \\
% Unit data multiplication cost, $\phi$ & 50 pJ/bit \\
% Upload cost of unit data from user $i$ to AP $m$ , $\varsigma_{i,m}$&  3 mJ/Mb \\
% CPU clock cycles on the edge server $k$,  $f_{k}$ & [2,10] GHz \\
% Upload cost of unit data between edge servers, $\varsigma_{k,l}$ & 5 mJ/Mb\\
% Bandwidth between user $i$ and AP $m$, $B_{im}$ & [20,50] Mbps\\
% Bandwidth between edge servers, $B_{kl}$ & 100 Mbps\\
% Learning rate of actor-critic network & 3e-4\\
% Reward discount, $\gamma$ & 0.99\\
% Soft update coefficient, $\tau$ & 0.01\\
% Replay buffer size & 1e5\\
% Experience mini batch size & 256\\
% \bottomrule
% \end{tabular*}
% \end{table}

\vspace{0em}
\subsection{Experiment settings.}
\textbf{Simulation settings}. The EC scenario is set on a 2000m $\times$ 2000m plane. The number of users is the same as the number of vertices in the graph dataset. The service scope of each edge server is set to 500m $\times$ 500m, so there will be four edge servers in the EC system. Furthermore, the positions of users and the edge devices are set randomly and only the user's position can dynamically change. In our EC system, the service capacity of edge servers will be randomly assigned, and the service capacity of the edge servers is categorized into three levels: high, medium, and low. The values corresponding to the three service capacity levels of edge servers are $\frac{5}{4}Mean$, $Mean$, and $\frac{3}{4}Mean$, where the value of $Mean$ is the quotient of the number of users and the number of edge servers. Lastly, the values for $B_{max1}, B_{max2}, P_{max1}$ and $P_{max2}$ are set as: 5000 MHz, 500 MHz, 1.5 W, and 60 mW, respectively. 

\begin{table}[width=\linewidth,cols=2,pos=h]
\footnotesize
\caption{Parameters of the system.}\label{table2}
\begin{tabular*}{\tblwidth}{@{} CC@{} }
\toprule
Parameters & Value\\
\midrule
Noise power, $\sigma^2$ & -110 dBm \\
Number of users, $N$ & [50,300] \\
Transmission power of user $i$, $P_{i}$ & [2,5] mW \\
Transmission power of edge server $k$, $P_{k}$ & [10,15] mW \\
Unit data aggregation cost of GNN inference & 20 pJ/bit \\
Unit data update cost of GNN inferenc, $\vartheta$ & 100 pJ/bit \\
Unit data multiplication cost, $\phi$ & 50 pJ/bit \\
Upload cost of unit data from user $i$ to AP $m$ , $\varsigma_{i,m}$&  3 mJ/Mb \\
CPU clock cycles on the edge server $k$,  $f_{k}$ & [2,10] GHz \\
Upload cost of unit data between edge servers, $\varsigma_{k,l}$ & 5 mJ/Mb\\
Bandwidth between user $i$ and AP $m$, $B_{im}$ & [20,50] MHz\\
Bandwidth between edge servers, $B_{kl}$ & 100 MHz\\
Learning rate of actor-critic network & 3e-4\\
Reward discount, $\gamma$ & 0.99\\
Soft update coefficient, $\tau$ & 0.01\\
Replay buffer size & 1e5\\
Experience mini batch size & 256\\
\bottomrule
\end{tabular*}
\end{table}

\textbf{Training settings.} 
All experiments were conducted on our lab server equipped with an Intel Core i7-14700k CPU and two NVIDIA GeForce RTX 4090 GPUs. The experimental code is based on Python 3.10, and the actor-critic networks are implemented by the network module of Pytorch 2.1.2. In addition, Four GNN models are used in our experiment which are GCN, GAT~\cite{li2024chatmdg}, GraphSAGE~\cite{hamilton2017inductive} and SGC~\cite{wu2019simplifying}. We use the instance from Pytorch Geometric 2.5.3 to implement Four GNN models. All GNN models are pre-trained and have an accuracy of between 60\% and 80\% for node classification tasks. In our experiment, the vertices in the graph dataset are one-to-one with the user, and the classification of the vertices is treated as the user's task. All networks in our simulation contain three layers and each layer has 64 neurons. To ensure sufficient exploration for DRLGO, we set its exploration rate to 0.1. The experimental parameter settings are based on those used in works~\cite{cao2024dependent} and~\cite{zhang2020energy}, with the specific parameter configurations shown in Table~\ref{table2}.

\begin{figure}[htbp]
    \centering
    \includegraphics[width=0.8\linewidth]{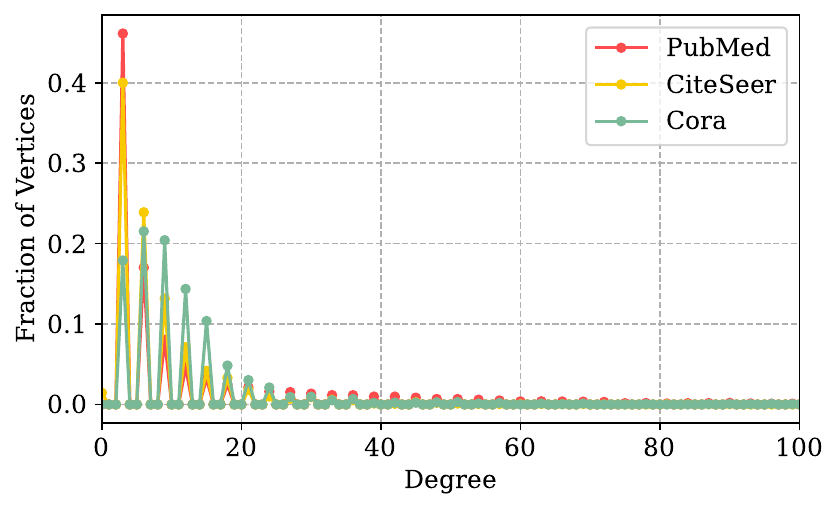}
    \caption{Vertices degree distribution for different datasets.}
    \label{dataset_degree_distribution}
\end{figure}
% The settings for other simulation parameters are shown in Table~\ref{table2}.
% 之前table2后面是没有结尾的点的

\textbf{Datasets in experiment.}
Three citation network datasets are considered in our experiments which are CiteSeer, Cora, and PubMed \cite{yang2016revisiting}, \cite{sen2008collective}. Each dataset only has one citation network, and contains bag-of-words representations of documents as well as citation links between documents. Moreover, each citation network is a graph in which documents serve as vertices in the graph and citation links act as edges in the graph. Next, the detailed information of datasets and how to use them in our EC system will be described. First, there are 3327 documents and 9104 citation links in CiteSeer, with documents organized into 6 categories. Second, there are 2708 documents and 10556 citation links in Cora,  and the documents are organized into seven categories. Last, there are 19717 documents and 88648 citation links in PubMed where documents are organized into 3 categories. The dimension size of the document feature vector in CiteSeer, Cora, and PubMed are 3703, 1433, and 500, respectively. The degree distribution of the vertices in three datasets is shown in Fig.~\ref{dataset_degree_distribution}. We randomly sample 300 documents and 4800 citation links from PubMed as a training dataset and resample dataset from three datasets at the evaluation phase.  In our EC system, the number of users is equal to the sampled documents, and the data features of each document are used as the user's task. Each dimension of the document data feature corresponds to a user data size of 1 kb and dimensions greater than 1500 are considered 1500. Finally, the tasks of users will be processed by GNN as classification tasks on the edge server.

% \vspace{-1em}
\textbf{Baselines.} Since the solution of the second subproblem depends on the result of the first one (as shown in line 8 of Algorithm~\ref{alg2}), the evaluation results of DRLGO already reflect the overall performance of our proposed solution. We compared DRLGO with the other three task offloading methods. They are as follows:

1) PPO-based task offloading method (\textit{PTOM}): PPO~\cite{schulman2017proximal} is a DRL method in which an agent uses the global environment state to explore task offloading strategies. The network architecture of \textit{PTOM} is the same as DRLGO, but it does not use the HiCut and subgraph offloading constraints.

2) Greedy method (\textit{GM}): \textit{GM} offloads user tasks to the nearest edge server.

3) Random method (\textit{RM}): \textit{RM} randomly offloads user tasks to any server without considering any scenario information.

The reason for choosing these comparative algorithms is that they have been widely used and studied in task offloading problems.

\begin{figure}[htbp]
  % \centering
  % 子图(a)
  \begin{subfigure}{0.49\linewidth}
    % \centering
    \includegraphics[width=\linewidth]{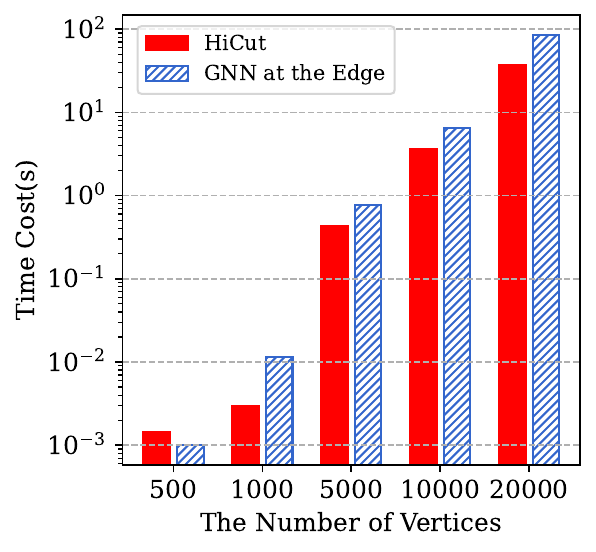} 
    \caption{Sparse graph.}
    \label{graph_cut:sparse}
  \end{subfigure}
  % \hfill
  % \hspace{-0.5em}
  % 子图(b)
  \begin{subfigure}{0.49\linewidth}
    % \centering
    \includegraphics[width=\linewidth]{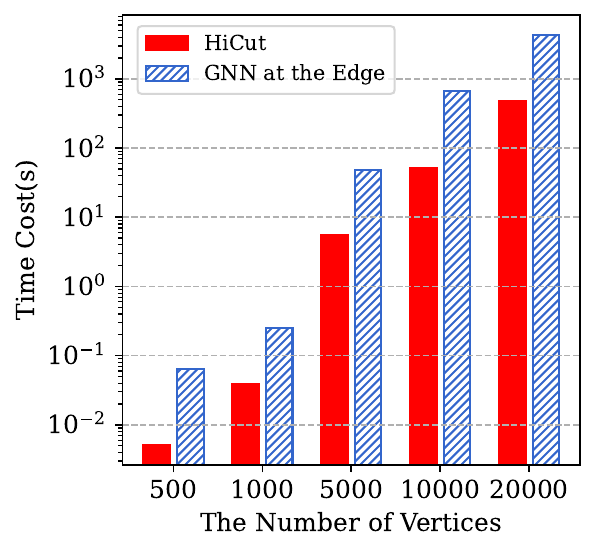} 
    \caption{Non-sparse graph.}
    \label{graph_cut:non_sparse}
  \end{subfigure}
  \caption{The comparison of different graph cut methods.}
  \label{graph_cut}
\end{figure}

\vspace{-0.5em}
\subsection{Graph segmentation performance}
In order to verify the graph cut performance of the proposed HiCut, we compare it with the graph cut method in the work \cite{zeng2022gnn} on graphs with different numbers of vertices and edges. Since the number of vertices and edges in our dataset is small, we randomly generated edges for different numbers of vertices and categorized them into sparse and non-sparse graphs based on the number of edges. The number of vertices and edges of the sparse and non-sparse graphs are set to:

1) \textbf{Sparse graph:} the number of vertices is 500 to 20000 and the number of edges is 5010 to 800040. 

2) \textbf{Non-sparse graph:} the number of vertices is 500 to 20000 and the number of edges is 500100 to 8000400. 

For the comparison algorithm, we follow the graph cut method from the work \cite{zeng2022gnn}, which performs graph cut operation iteratively. The number of iterations depends on the number of edge servers because it selects a pair of edge servers as the source point and the sink point for each iteration, and the processing involves the vertices and edges between these two servers. The edge weights in the graph are set to random integers between 1 and 100 and the number of edge servers is 25.
Experimental results in Fig.~\ref{graph_cut} show that the proposed HiCut outperforms the comparison method, especially when there are more edges in the graph. In Fig.~\ref{graph_cut:sparse}, it can be seen that when the number of vertices is 500, HiCut takes a little more time cost than the comparison method, this is because when the comparison algorithm performs graph cut operation between pairs of edge servers, the number of vertices and edges is much smaller than 500 and 5010. Furthermore, we can observe from the results that as the number of nodes and edges increases, the advantage of HiCut becomes more obvious. As shown in Fig.~\ref{graph_cut:non_sparse}, HiCut is almost an order of magnitude faster than the comparison method in the processing of non-sparse graphs. This is because the complexity of HiCut is $O(V^2 + |VE|)$ while the complexity of the comparison algorithm is $O(V^2E)$.

% CiteSeer
\begin{figure*}[htbp]
\centering
\begin{subfigure}{0.24\linewidth}
\includegraphics[width=\linewidth]{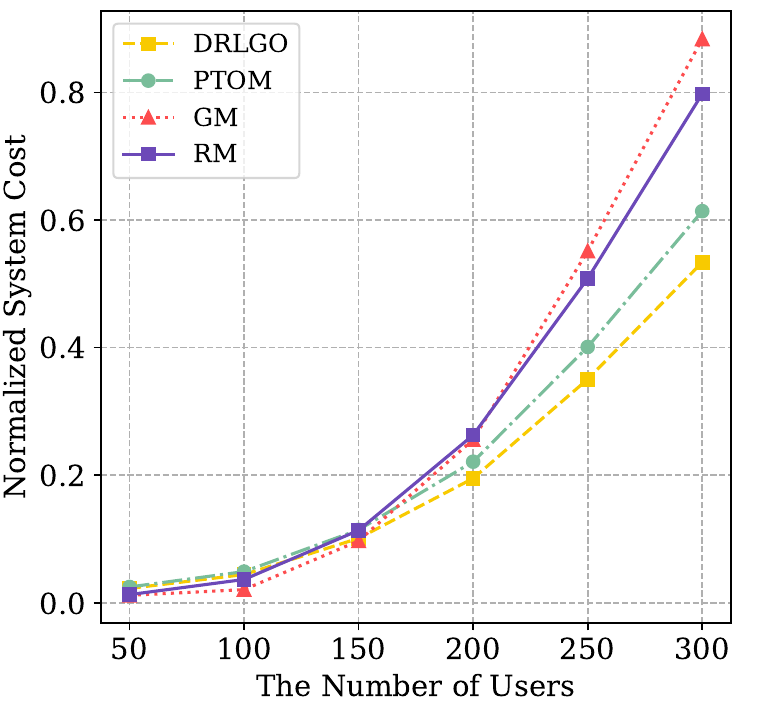}
\caption{The change of user's \\ number.}
\label{CiteSeer:number}
\end{subfigure}%
\hfill
\begin{subfigure}{0.24\linewidth}
\includegraphics[width=\linewidth]{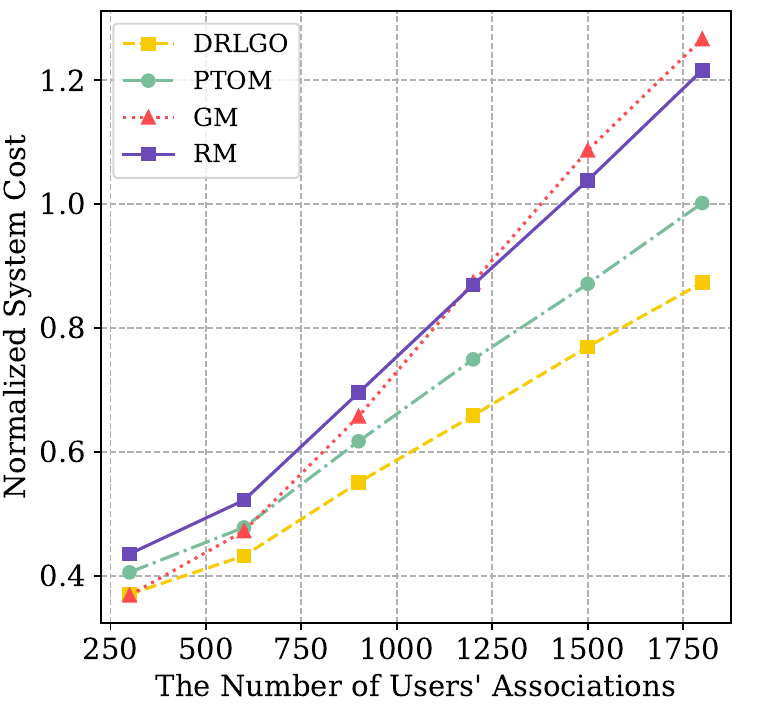}
\caption{The change of user's associations.}
\label{CiteSeer:association}
\end{subfigure}%
\hfill
\begin{subfigure}{0.24\linewidth}
\includegraphics[width=\linewidth]{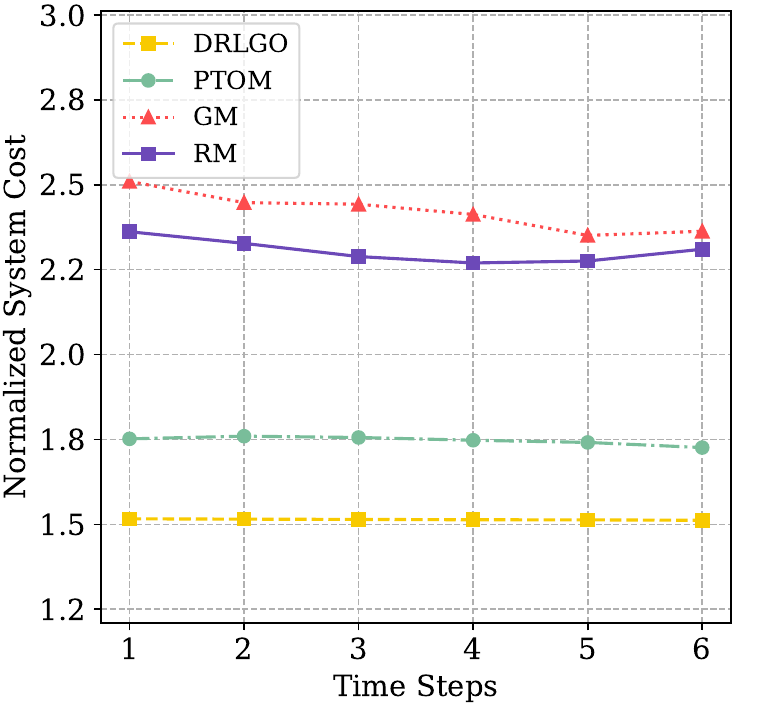}
\caption{The change of user's positions.}
\label{CiteSeer:position}
\end{subfigure}%
\hfill
\begin{subfigure}{0.24\linewidth}
\includegraphics[width=\linewidth]{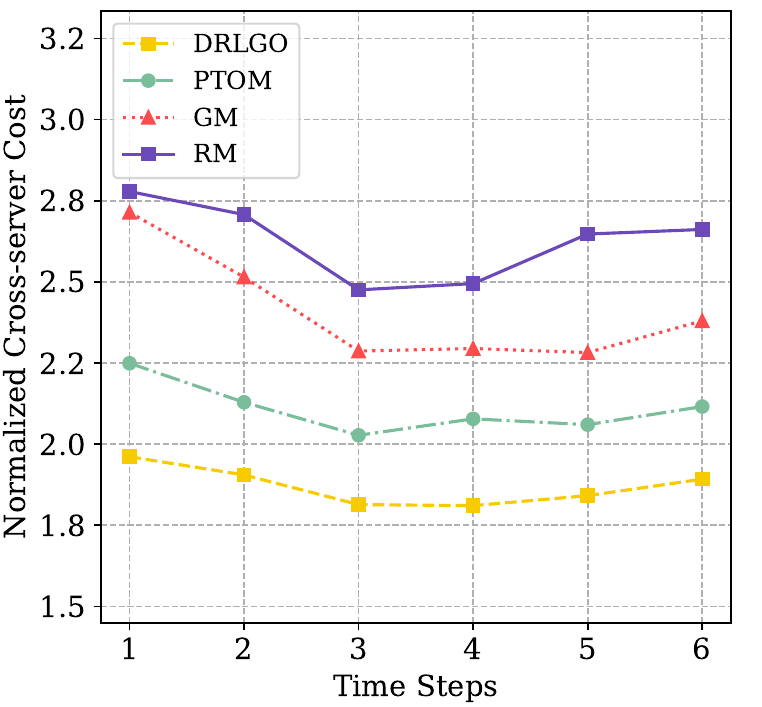}
\caption{The comparison of cross-server cost.}
\label{CiteSeer:result}
\end{subfigure}
\caption{Dynamic performance of different methods on CiteSeer.}
\label{CiteSeer}
\end{figure*}

% Cora
\begin{figure*}[htbp]
\centering
\begin{subfigure}{0.24\linewidth}
\includegraphics[width=\linewidth]{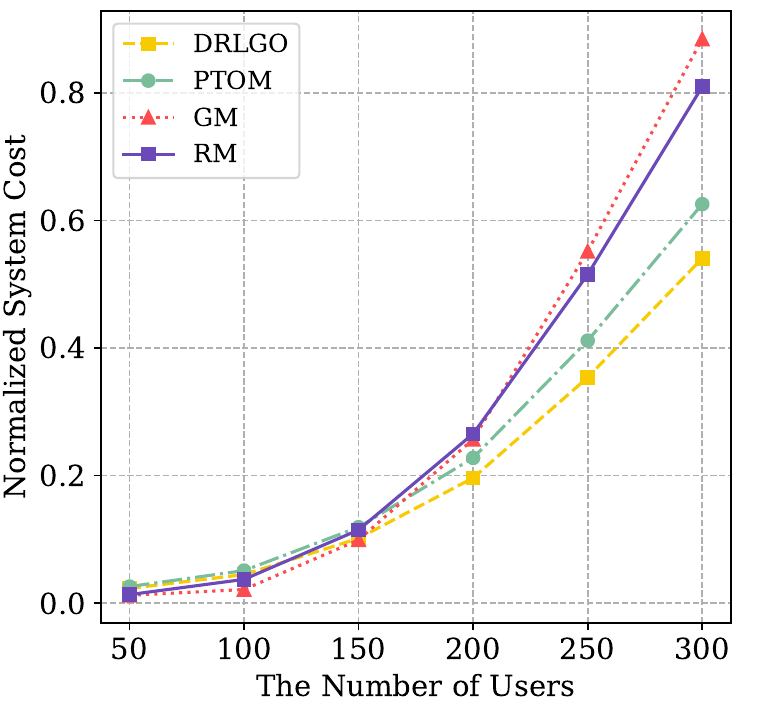}
\caption{The change of user's \\ number.}
\label{Cora:number}
\end{subfigure}%
\hfill
\begin{subfigure}{0.24\linewidth}
\includegraphics[width=\linewidth]{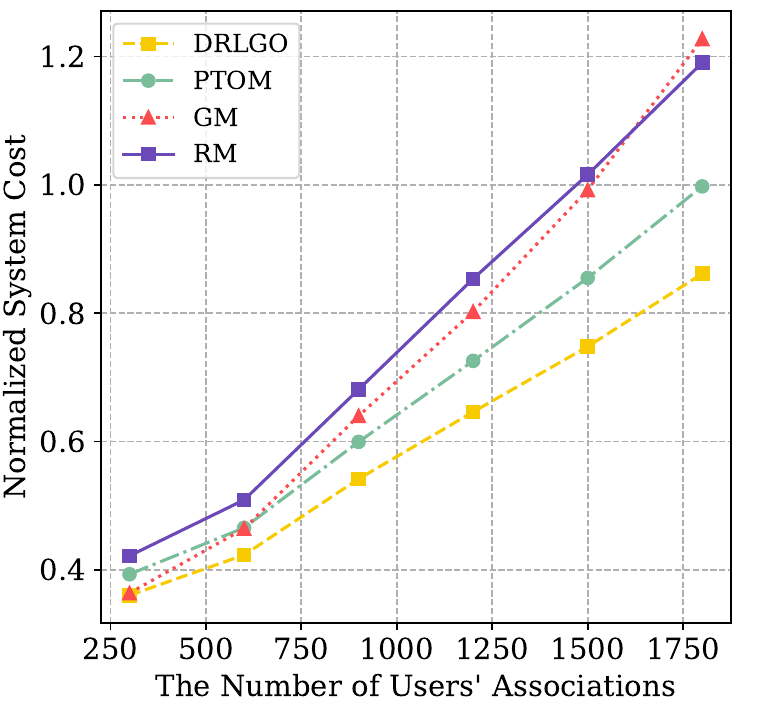}
\caption{The change of user's associations.}
\label{Cora:association}
\end{subfigure}%
\hfill
\begin{subfigure}{0.24\linewidth}
\includegraphics[width=\linewidth]{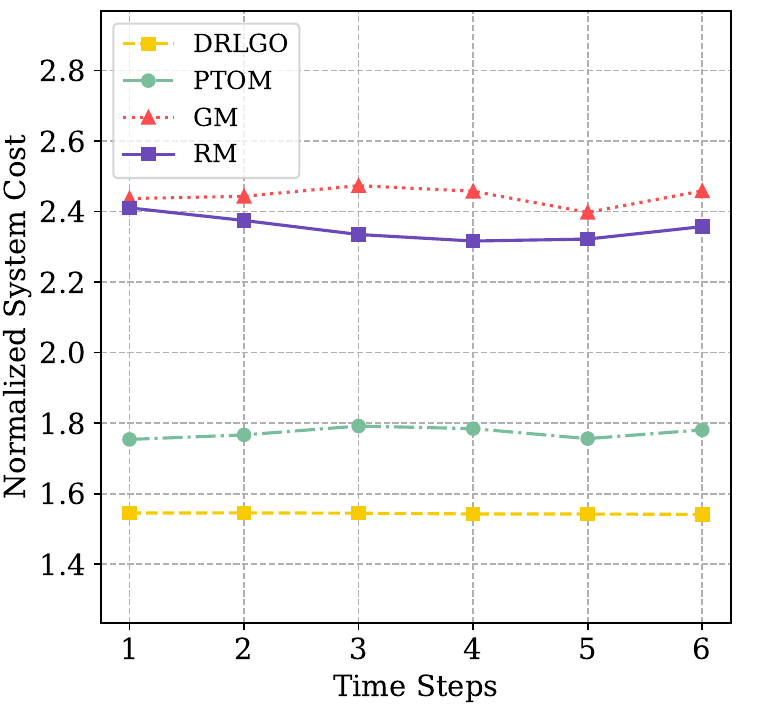}
\caption{The change of user's \\ positions.}
\label{Cora:position}
\end{subfigure}%
\hfill
\begin{subfigure}{0.24\linewidth}
\includegraphics[width=\linewidth]{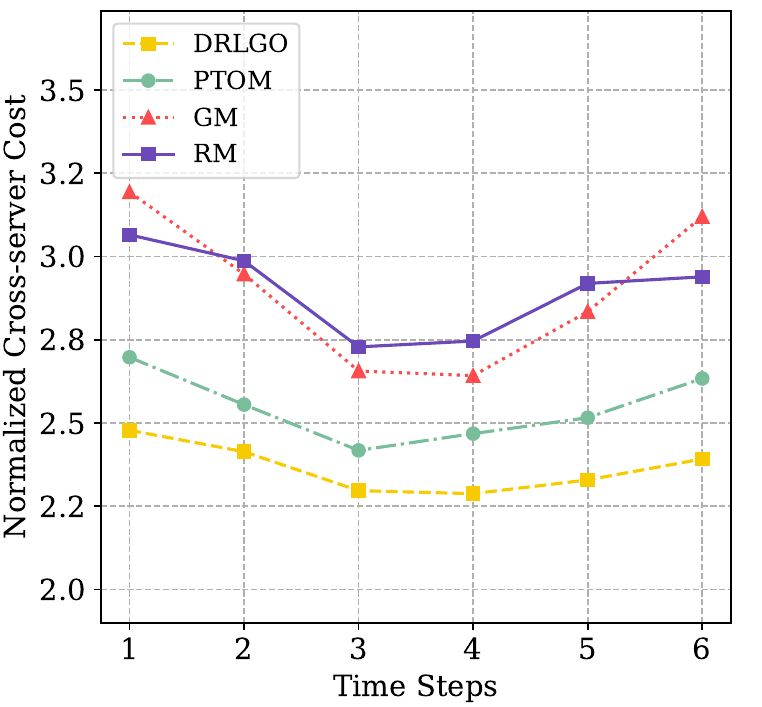}
\caption{The comparison of cross- \\server cost.}
\label{Cora:result}
\end{subfigure}
\caption{Dynamic performance of different methods on Cora.}
\label{Cora}
\end{figure*}

\subsection{Performance comparison}
In this section, we evaluate the performance of DRLGO and baseline methods across different datasets and GNN models. In simulation experiments, we first compare the system costs of all methods under dynamic user states. The dynamic states of users are achieved by changing the number of users, the locations of users, and the associations of users. Next, we randomly change user states at different time steps and compare the cross-server communication cost of various methods. Finally, we evaluate the performance of all methods using different GNN models.

\begin{figure*}[htbp]
\centering
\begin{subfigure}{0.24\linewidth}
\includegraphics[width=\linewidth]{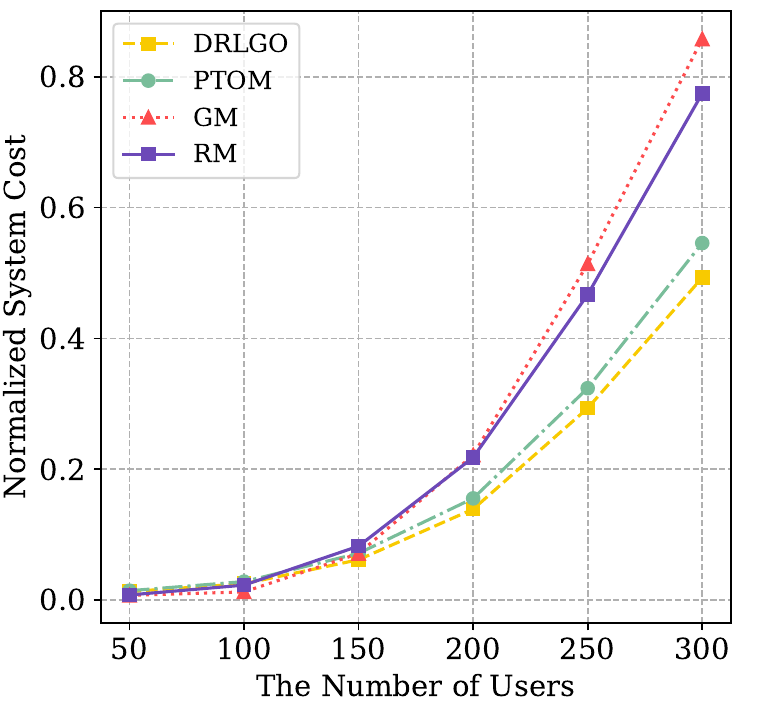}
\caption{The change of user's \\ number.}
\label{PubMed:number}
\end{subfigure}%
\hfill
\begin{subfigure}{0.24\linewidth}
\includegraphics[width=\linewidth]{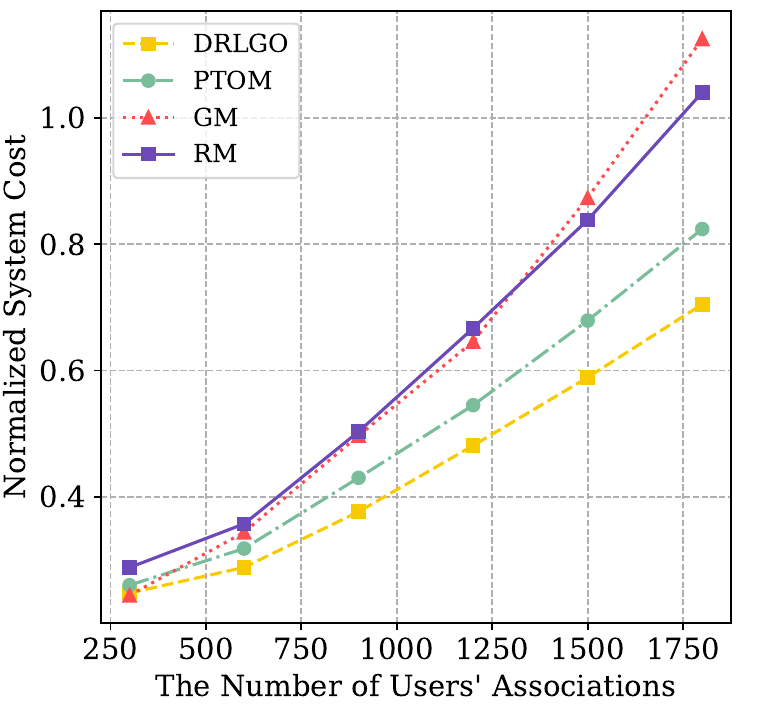}
\caption{The change of user's associations.}
\label{PubMed:association}
\end{subfigure}%
\hfill
\begin{subfigure}{0.24\linewidth}
\includegraphics[width=\linewidth]{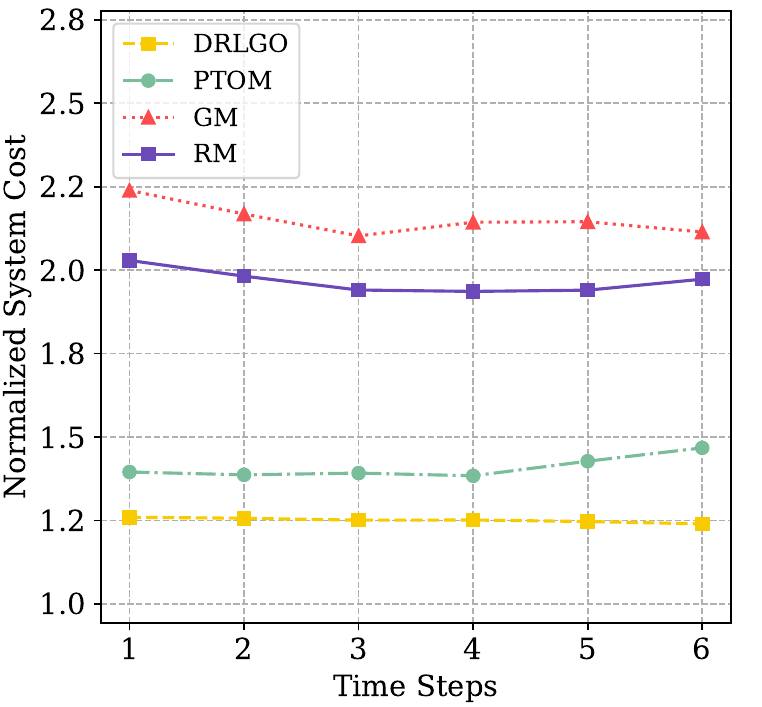}
\caption{The change of user's positions.}
\label{PubMed:position}
\end{subfigure}%
\hfill
\begin{subfigure}{0.24\linewidth}
\includegraphics[width=\linewidth]{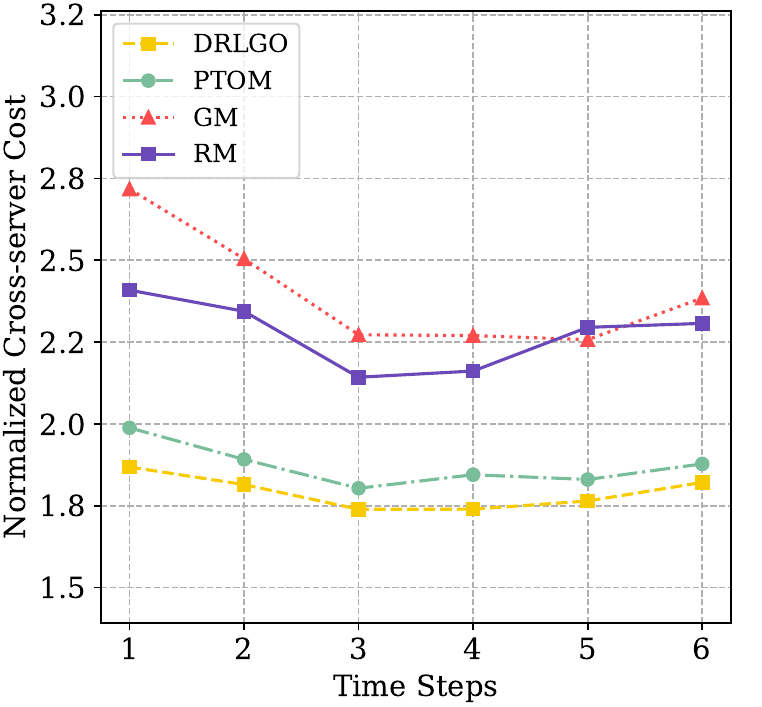}
\caption{The comparison of cross-server cost.}
\label{PubMed:result}
\end{subfigure}
\caption{Dynamic performance of different methods on PubMed.}
\label{PubMed}
\end{figure*}

First, we evaluate and compare the system cost of various methods under dynamic user states in different datasets. In these evaluations, the used GNN model is GCN, the number of users changed from 50 to 300, the number of associations between users changed from 300 to 1800 and the positions of users changed randomly. Each method was evaluated 10 times, and the average value was taken as the result. From Fig.~\ref{CiteSeer} to Fig.~\ref{PubMed}, it can be seen that as the number of users and their associations increase, the system costs for all methods also rise. This is because the increase in user tasks and user associations requires more energy and time cost of the EC system. It is worth noting that the proposed DRLGO has better performance than the other methods in the user dynamic state. This is due to the fact that DRLGO is trained based on the optimized graph layout obtained by HiCut, which takes into account the system cost of user task uploading and task processing and minimizes cross-server communication cost by offloading user tasks from the same subgraph to the same edge server as much as possible. The reason why the performance of \textit{PTOM} is inferior to DRLGO is that the latter cannot adequately take into account the communication cost caused by GNN inference. Meanwhile, \textit{GM} and \textit{RM} merely consider offloading user tasks to the nearest edge server or randomly to any edge server, so they perform poorly than \textit{PTOM} and DRLGO as the number of users and associations increase. However, \textit{RM} may be slightly better than \textit{GM}, due to the fact that \textit{RM} may offload a user task with its neighboring tasks to the same edge server or offload the user task to the nearest server.

% PubMed

\begin{figure}
\centering
\begin{subfigure}{\linewidth}
\includegraphics[width=\linewidth]{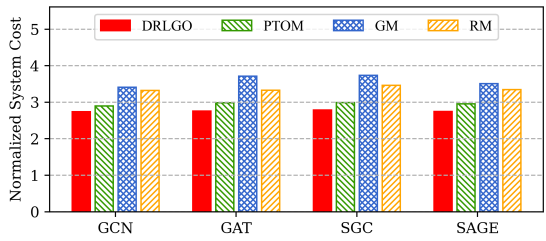}
\caption{The system cost on CiteSeer.}
\label{GNN_model:CiteSeer}
\end{subfigure}%
\hfill
\begin{subfigure}{\linewidth}
\includegraphics[width=\linewidth]{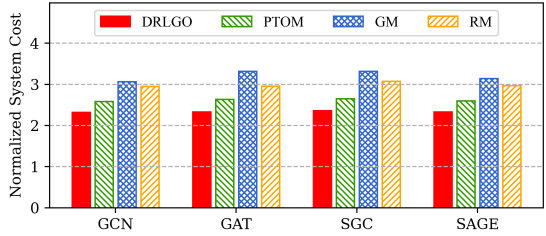}
\caption{The system cost on Cora.}
\label{GNN_model:Cora}
\end{subfigure}%
\hfill
\begin{subfigure}{\linewidth}
\includegraphics[width=\linewidth]{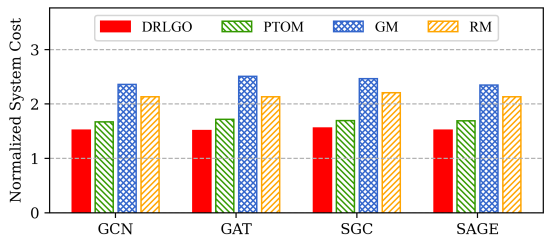}
\caption{The system cost on PubMed.}
\label{GNN_model:PubMed}
\end{subfigure}
\caption{The system cost of various methods on different datasets using different GNN models.}
\label{GNN_model}
\end{figure}

Next, we investigate the performance of each method on different datasets by randomly changing the position of all users at each time step. The experimental results are shown in Fig.~\ref{CiteSeer:position}, Fig.~\ref{Cora:position} and Fig.~\ref{PubMed:position}, and it can be seen that the proposed DRLGO is better than the other methods overall. Moreover, when facing user mobility,  DRLGO has smaller system cost fluctuation compared to the \textit{PTOM}. The remaining two methods perform poorly by one-sidedly considering the distance factor or randomly offloading tasks, and \textit{RM} has the most fluctuating system cost. To evaluate the effectiveness of DRLGO in minimizing cross-server communication caused by GNN inference, we randomly change user states across multiple time steps and record the cross-server communication cost for each method. Each method was evaluated 10 times, and the average cross-server communication cost was taken as the result. The GNN model used in the experiment is GCN, and the experimental results are shown in Fig.~\ref{CiteSeer:result}, Fig.~\ref{Cora:result} and Fig.~\ref{PubMed:result}. The experimental results show that the proposed DRLGO achieves the best performance across different datasets, thanks to the graph layout optimization by HiCut based on GNN aggregation characteristics, DRLGO can adapt to different datasets and explore the optimal graph offloading strategy.

Finally, to investigate the performance of different methods on different GNN models, we set the number of users to 300 and user associations to 4800. Each method is evaluated 10 times on different GNN models, and the average results are taken as the final performance. The comparison results of each method on different datasets are shown in Fig.~\ref{GNN_model}. It can be seen that although the system cost of each method fluctuates on different datasets due to the differences in user task sizes and user associations, the proposed DRLGO still maintains the minimum system cost. The reason is that, although the GNN models are different, they all require aggregation operations when handling graph tasks. Our proposed solution optimizes the graph layout based on the aggregation characteristics of GNNs to reduce communication between servers. It also demonstrates the strong model adaptation capability of DRLGO.

\subsection{Convergence performance}
\begin{figure}[htbp]
    \centering \includegraphics[width=0.8\linewidth,height=4cm]{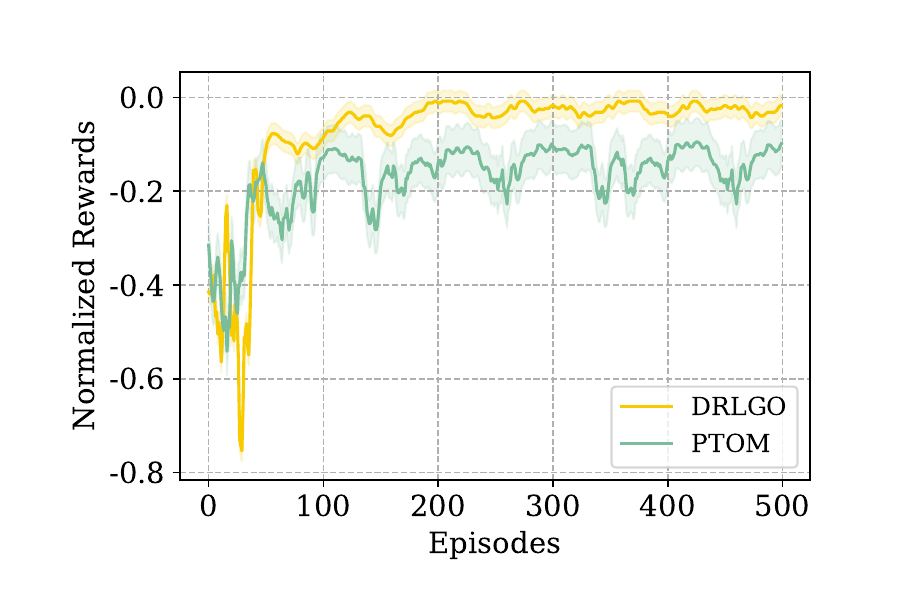}
    \caption{The comparison of cost convergence between DRLGO and PTOM during training.}
    \label{convergence_res}
\end{figure}
The reward curves of DRLGO and \textit{PTOM} during the training process are shown in Fig.~\ref{convergence_res}. During the training process we take the negative value of the system cost as the reward value, 300 documents are randomly sampled from the dataset as training data, which represent the users in the EC scenario. The number of associations between the users is set to 4800 and the associations of users will be randomly generated for less than 4800. Each round of training will dynamically change the user states with a 20\% change rate in both the number of users and their associations. All user tasks being offloaded and processed represent the completion of a training round. Since the training data is randomly sampled and dynamically simulates the possible states of the user, the trained model can be applied to different datasets. From Fig.~\ref{convergence_res}, it can be seen that the proposed DRLGO can obtain more stable and higher reward values, which is due to the fact that multiple agents of the DRLGO can not only give actions independently with the local observation during the training but also adjust their own strategies according to the actions and rewards of other agents. Meanwhile, the constraints on subgraph offloading in the reward function of DRLGO minimize the cross-server communication cost during GNN inference. This ensures that DRLGO maintains stable reward values even under dynamic user states. In contrast, \textit{PTOM} only has one agent to explore offloading strategies during training, without any constraints on cross-server communication cost. As a result, even if the model converges, there are still obvious fluctuations in reward values under dynamic user states.

\subsection{Ablation study}

\begin{figure}[htbp]
    \centering
    \includegraphics[width=0.9\linewidth]{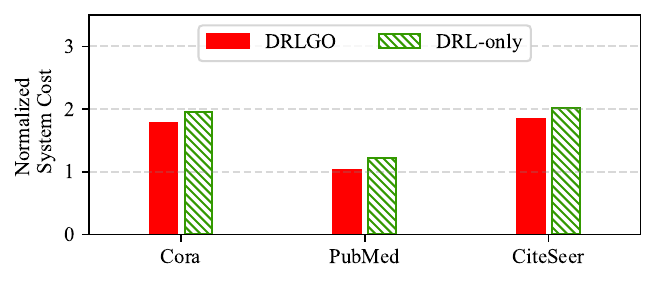}
    \caption{The comparison of system cost between DRLGO and DRL-only methods.}
    \label{ablation_res}
\end{figure}

To validate the effectiveness of HiCut in minimizing cross-server communication cost, we compared the system cost of DRLGO with and without HiCut, and the latter is called \textit{DRL-only}, which is trained without the reward function constraint of subgraph offloading, and just uses MADDPG as the offloading algorithm. DRLGO and \textit{DRL-only} were evaluated 10 times at different time steps. The number of users is set to 300 and the number of user associations is set to 4800. Fig.~\ref{ablation_res} shows the system energy cost of DRLGO compared to \textit{DRL-only} on different datasets. The experimental results show that DRLGO outperforms DRL-only. This is because DRLGO uses HiCut to optimize the original graph layout and exploring the graph offloading strategy based on optimized graph layout can effectively reduce the cross-server communication during GNN inference.

\section{Conclusion}
\label{Conclusion}
This paper investigates the problem of minimizing the cost of user task offloading and processing in GNN-based EC systems. To address this problem, we propose a new GNN-based EC architecture called GraphEdge. 
% This architecture first perceives and represents the data association of the users in the EC system as a graph, which is the original graph layout. Then, to minimize the large cross-server communication cost caused by GNN inference, we propose the graph layout optimization algorithm, which optimizes the original graph layout for the aggregation characteristics of GNN. And the optimized graph layout contains a series of weakly associated subgraphs and task offloading based on these subgraphs can minimize the cost of cross-server communication. In addition, in order to explore the optimal graph offloading strategy based on the optimized graph layout, we propose a DRL-based graph offloading algorithm, which gives optimal graph offloading strategy based on the subgraphs of the new graph layout, it determines which edge server the user tasks in different subgraphs are offloaded to. Experimental results show that proposed graph offloading algorithm can dynamically learn the complex graph layout and scenario information adequately. It also performs optimally on different GNN models and dynamic environments. Finally, as the part of our architecture, the good performance orevisedf the graph layout optimization algorithm and DRL-based graph offloading algorithm also proves the effectiveness of our proposed architecture.
First, the data associations between users in the EC system will be perceived and represented by this architecture as the original graph layout. Then, due to the significant cross-server communication cost caused by GNN inference, we propose a graph layout optimization algorithm for this problem based on the aggregation characteristics of GNN. After the optimization of original graph layout, multiple weakly associated subgraphs are obtained and during GNN inference, the communication cost between edge servers is minimized. Furthermore, since our goal is to minimize the EC system cost, we propose a DRL-based graph offloading algorithm based on the optimized graph layout. This algorithm comprehensively considers various costs of the EC system and dynamically provides the optimal graph offloading strategy based on the optimized graph layout. Experimental results show that the proposed graph layout optimization algorithm can minimize the communication cost between edge servers.  Meanwhile, the graph offloading algorithm can dynamically and adequately learn the complex graph layout and scene information, it has good model and environment adaptability, and can dynamically provide the optimal task offloading strategy to achieve the minimized system cost. Finally, as the part of our architecture, the good performance of the proposed algorithms also demonstrates the effectiveness of the proposed architecture.
In future work, we will consider the dynamic edge server states, such as using UAVs and smart vehicles as mobile edge servers to provide GNN computation services. Similarly, our work will continue to focus on optimizing the cost of the EC system.

\section*{Acknowledgement}
\label{Acknowledgement}

This work was supported in part by the National Natural
Science Foundation of China (62462002) and partially sup-
ported by the Natural Science Foundation of Guangxi, China (Nos. 2025GXNSFBA069394, 2025GXNSFAA069958).

\printcredits
% \bibliographystyle{cas-model2-names}
\bibliographystyle{unsrt}
% \bibliography{reference}

\end{document}